\documentclass[10pt,twocolumn,letterpaper]{article}

\usepackage{cvpr} 

\usepackage{graphicx}
\usepackage{dblfloatfix} 
\graphicspath{{figures/}}
\usepackage{booktabs}
\usepackage{url}
\usepackage{subcaption}
\usepackage{dblfloatfix}
\usepackage{multirow}
\usepackage{tabularx}
\usepackage{array}
\usepackage{comment}

\newcolumntype{Y}{>{\raggedright\arraybackslash}X}

\definecolor{cvprblue}{rgb}{0.21,0.49,0.74}
\usepackage[pagebackref,breaklinks,colorlinks,allcolors=cvprblue]{hyperref}

\def\paperID{*****} 
\def\confName{CVPR}
\def\confYear{2026}

\title{SPEAR-NeXT: Causal Latent Forecasting Across Multiple Horizons for Spectral–Temporal Earth Representation Learning}

\author{
Rajiv Ranjan \quad Udaiveer Singh \quad Shashank Tamaskar\\
Plaksha University\\
Mohali, Punjab, India\\
{\tt\small \{rajiv.ranjan, udaiveer.singh.ug23, shashank.tamaskar\}@plaksha.edu.in}
\and
Dharmendra Saraswat\\
Purdue University\\
West Lafayette, Indiana, USA\\
{\tt\small saraswat@purdue.edu}
}

\begin{document}
\maketitle
\begin{abstract}

Earth observation is inherently dynamic, yet temporal information in many
foundation models is learned through reconstruction, invariance, or
retrospective sequence summarization. \textbf{SPEAR-NeXT} is introduced as a
compact pixel-wise multimodal spectral--temporal foundation model in which
temporal self-supervision is formulated as past-only, multi-horizon latent
Earth-state prediction. Instantaneous states are first encoded by the
pretrained SPEAR model from optical, radar, and environmental observations
into compact 32-dimensional embeddings. Their temporal evolution is then
modeled by a causally masked Transformer that predicts multiple future latent
states from preceding observations. Relative temporal order is represented
using Rotary Position Embeddings, while month and year embeddings encode
seasonal phase and inter-annual context.

Horizon-weighted latent trajectory supervision is used to capture local
continuity, intermediate state transitions, and longer-term seasonal
structure. Cosine alignment preserves latent-space direction, while latent
regression constrains coordinate scale and limits magnitude drift. Separate
regional models are pretrained over India and the contiguous United States
using monthly observations from 2020--2024. The learned representations are
evaluated through frozen and lightweight adaptation on land-cover mapping,
crop classification, phenological estimation, and crop-yield forecasting.
SPEAR-NeXT achieves land-cover classification accuracies of $94.81\%$ in
India and $88.78\%$ in CONUS, obtains the lowest errors for sowing-date,
harvest-date, and yield estimation on SICKLE, and reaches a mean $R^2$ of
$0.721$ for USDA-NASS crop-yield prediction, compared with $0.644$ for
TESSERA and $0.613$ for Presto. Prediction-head and kriging ablations further
indicate that the gains are primarily associated with the pretrained temporal
representation rather than downstream capacity or spatial post-processing.
These results support causal multi-horizon latent forecasting as an effective
objective for compact and reusable pixel-level Earth representations.

\end{abstract}    
\section{Introduction}
\label{sec:intro}

Earth observation (EO) is not a collection of independent images, but a record of a continuously changing surface. Vegetation growth, soil-moisture variation, inundation, disturbance, recovery, and land management unfold through trajectories whose meaning is often ambiguous in a single observation. Temporal structure is therefore central to land-cover monitoring, change analysis, crop-type mapping, phenological assessment, and environmental forecasting \cite{Cong_2022_NeurIPS,Tseng_2023_Presto}.

Self-supervised learning has made it possible to exploit large unlabeled satellite archives and has driven rapid progress in EO foundation models. Existing approaches reconstruct masked spatial, spectral, or temporal content, align observations across sensors or augmentations, or summarize multi-date inputs into transferable embeddings. SatMAE extends masked autoencoding to temporal and multispectral imagery, CROMA combines radar--optical contrastive learning with masked reconstruction, Presto reconstructs structured pixel-level sensor time series, AnySat uses joint-embedding prediction across heterogeneous EO inputs, and AlphaEarth Foundations generates time-conditioned embedding fields from multisource observations \cite{Cong_2022_NeurIPS,Fuller_2023_NeurIPS,Tseng_2023_Presto,Astruc_2025_CVPR,Brown_2025_AlphaEarth}. These models demonstrate that temporal and multisensor context are valuable. The remaining gap is therefore not the absence of time, but the limited use of \emph{directional predictive supervision}: most temporal objectives recover missing observations, align alternative views, or summarize an observed interval, rather than requiring a representation at time $t$ to predict several unseen states after $t$ using only preceding context.

This distinction is important for pixel-level satellite time series. At the pixel scale, spatial context is deliberately minimized, and temporal evolution becomes a primary source of information. A pixel's spectral response may change gradually because of vegetation development, surface moisture, disturbance, management, or seasonal climate. A useful temporal representation should consequently encode not only what has already been observed, but also which components of the current state remain informative about its future trajectory. We therefore ask:

\begin{quote}
\emph{What information must a compact pixel representation preserve to remain predictive across multiple future horizons?}
\end{quote}

\begin{quote}
\emph{SPEAR: What is the latent state now?
SPEAR-NeXT: How will that latent state evolve?
}
\end{quote}

Predictive representation learning provides a promising alternative to observation reconstruction. I-JEPA predicts latent target-region representations from visible image context, and V-JEPA extends feature prediction to video without reconstructing pixels \cite{Assran_2023_CVPR,Bardes_2024_TMLR}. In EO, future-image forecasting has also been formulated directly in observation space, for example by predicting future Sentinel-2 imagery conditioned on environmental variables \cite{RequenaMesa_2021_CVPR}. These directions motivate prediction as a learning signal, but they leave open a distinct EO problem: causal prediction of multiple future \emph{pixel-level spectral states} for representation learning. Forecasting in latent space does not make the targets intrinsically noise-free; rather, it shifts the objective from reproducing every radiometric detail toward predicting the temporal structure retained by a pretrained encoder.

To address this problem, SPEAR-NeXT is introduced as a compact pixel-wise spectral--temporal foundation model that formulates temporal pretraining as causal latent Earth-state prediction. Here, \emph{causal} denotes autoregressive information flow: the prediction at time $t$ is computed only from states observed at or before $t$, and does not imply intervention-based causal inference. Similarly, an \emph{Earth state} refers to a learned spectral representation of a pixel rather than a complete physical state of the land surface.

SPEAR-NeXT follows a spectral-to-temporal decomposition. The previously published SPEAR encoder first maps each multispectral pixel into a compact latent state using reflectance together with continuous band-center wavelength and bandwidth metadata \cite{Ranjan_2026_WACV}. The temporal component then learns a transition operator over the resulting state sequence. In this decomposition, SPEAR estimates the instantaneous spectral state, whereas SPEAR-NeXT learns how that state evolves. The two-stage design also enables modular pretraining: spectral states can be computed once, cached, and subsequently used to train the temporal model without repeatedly processing raw multispectral measurements.

The temporal model predicts several future states directly from a causal Transformer representation. Two forms of temporal information are modeled separately. Rotary Position Embeddings (RoPE) encode sequence order and relative displacement within self-attention \cite{Su_2024_RoFormer}, while learnable month and year embeddings provide absolute seasonal phase and inter-annual context. This separation is useful because identical sequence offsets may occur at different points of an annual cycle, whereas the same calendar month can exhibit different conditions across years.

A central element of SPEAR-NeXT is horizon-weighted latent trajectory supervision. Multi-horizon forecasting is used not only to produce future predictions, but also to constrain what the representation preserves. A one-step objective can favor local continuity; supervision at multiple horizons additionally exposes the model to intermediate transitions and longer-range recurrence. Horizon-dependent weighting stabilizes near-term learning while retaining supervision from more distant states. Predicted and target trajectories are compared through a composite objective that combines cosine alignment with latent vector regression, thereby constraining both orientation and scale in the frozen SPEAR embedding space.

Although agriculture provides a demanding test because crop identity, phenological stage, management, and productivity depend strongly on temporal progression, SPEAR-NeXT is not formulated as an agriculture-specific model. Its pretraining objective is task-agnostic and is applied to diverse unlabeled land-surface observations. Evaluation is therefore conducted at two levels: embedding-space diagnostics measure horizon-wise predictability, magnitude drift, temporal self-similarity, and prediction--target alignment; downstream experiments measure transfer to general land-cover mapping and temporally demanding agricultural tasks using frozen or lightweight adaptation. Capacity-matched and prediction-head ablations are used to distinguish gains from temporal pretraining from those caused only by additional downstream parameters.

\paragraph{Contributions.}
The main contributions are:

\begin{itemize}
    \item A self-supervised temporal pretraining formulation for EO in which compact pixel-level spectral states are predicted causally across multiple future horizons, rather than reconstructed from bidirectional context or summarized retrospectively.

    \item A modular spectral-to-temporal decomposition that builds on the wavelength-aware SPEAR encoder and learns a separate causal transition model over its compact latent states.

    \item A horizon-weighted latent trajectory objective, together with complementary relative and calendar-time conditioning, that constrains the representation across local continuity, intermediate transitions, and longer-range temporal structure.

    \item A representation-focused evaluation protocol combining forecasting baselines, latent-geometry diagnostics, temporal and objective ablations, and capacity-matched downstream transfer on land-cover and agricultural tasks.
\end{itemize}

SPEAR-NeXT thus moves pixel-level temporal EO pretraining from encoding an already observed sequence toward learning representations that remain informative about how the sequence can evolve.

\section{Related Work}
\label{sec:related_work}

\subsection{Self-Supervised Learning and Earth-Observation Foundation Models}

The scale of modern satellite archives, together with the scarcity and uneven geographic distribution of ground-truth labels, has made self-supervised learning central to Earth-observation (EO) representation learning. In general computer vision, masked autoencoding and self-distillation have demonstrated that large models can learn reusable representations from unlabeled imagery and can subsequently support strong frozen-backbone transfer \cite{he2022mae,simeoni2025dinov3}. EO foundation models extend this paradigm by incorporating properties that are uncommon in natural imagery, including multispectral measurements, repeated observations, sensor heterogeneity, geographic scale, and irregular data availability.

Masked reconstruction remains the dominant EO pretraining objective. SatMAE extends masked autoencoding to temporal and multispectral satellite imagery through temporal embeddings, independent masking across acquisition times, and spectral group embeddings \cite{Cong_2022_NeurIPS}. Scale-MAE incorporates the ground sampling scale into positional encoding and reconstructs frequency components at multiple scales \cite{reed2023scalemae}, while Prithvi pretrains a multi-temporal Transformer on Harmonized Landsat--Sentinel-2 imagery for transfer to several EO tasks \cite{jakubik2023prithvi}. CROMA combines radar--optical contrastive alignment with masked reconstruction \cite{Fuller_2023_NeurIPS}, and SkySense learns multimodal spatiotemporal representations using multi-granularity contrastive and geographic prototype objectives \cite{guo2024skysense}. More recently, AnySat adopts a joint-embedding predictive architecture to support heterogeneous resolutions, scales, and modalities \cite{Astruc_2025_CVPR}, whereas AlphaEarth Foundations produces compact embedding fields by assimilating spatial, temporal, and measurement context across multiple sources \cite{Brown_2025_AlphaEarth}. Collectively, these models establish the value of large-scale EO pretraining; however, their temporal learning signals are primarily based on masked recovery, cross-view alignment, or conditional summarization rather than explicit causal prediction of future latent Earth states.

\subsection{Spectral and Sensor-Aware Representation Learning}

The spectral axis is physically structured: each channel measures radiance over a sensor-specific wavelength response rather than acting as an interchangeable image channel. SpectralGPT and S2MAE exploit this structure through spatial--spectral tokenization and masked reconstruction \cite{hong2024spectralgpt,li2024s2mae}. DOFA goes further by conditioning dynamic patch-embedding weights on channel wavelengths, allowing a shared Transformer to process different sensor configurations \cite{xiong2024dofa}. These methods substantially improve spectral and cross-sensor transfer, but they generally learn spectral structure jointly with spatial patches or cubes and optimize reconstruction at the observation level.

SPEAR-NeXT follows a complementary factorization. Its spectral encoder operates on the multispectral vector of an individual pixel and uses continuous band-center wavelength and bandwidth metadata to produce a compact latent state. This design treats wavelength metadata as a physical index of the measurement process, while avoiding the stronger claim that the representation is explicitly constrained by a physical process model. The resulting spectral state becomes the input to a separate temporal learner. This spectral-to-temporal decomposition distinguishes instantaneous state estimation from state evolution and enables the temporal objective to operate entirely in a compact embedding space.

\subsection{Temporal Modeling and Self-Supervision for Satellite Time Series}

Early satellite image time-series models used recurrent networks to summarize vegetation and land-cover trajectories \cite{russwurm2018recurrent}. Temporal convolutional networks subsequently improved parallelism and provided effective local-to-intermediate temporal receptive fields \cite{pelletier2019temporal}. Transformer-based models replaced recurrence with attention, enabling direct interaction between distant observations. The Pixel-Set Encoder with Temporal Attention Encoder demonstrated the effectiveness of temporal self-attention for parcel-level satellite time-series classification \cite{garnot2020satellite}, while thermal positional encoding showed that calendar-independent phenological coordinates can improve cross-region crop classification \cite{nyborg2022thermal}. These approaches established the importance of temporal order, acquisition timing, and seasonal phase, but they are principally trained for supervised classification and therefore learn temporal features that are coupled to a specific label space.

Temporal self-supervision has since become more prominent. SatMAE reconstructs temporally masked image patches \cite{Cong_2022_NeurIPS}, and Presto uses structured masking to reconstruct missing time points and sensor groups from pixel-level multisensor time series \cite{Tseng_2023_Presto}. Prithvi and SkySense also ingest multi-temporal imagery, while AlphaEarth learns time-conditioned summaries that support mapping at specified periods \cite{jakubik2023prithvi,guo2024skysense,Brown_2025_AlphaEarth}. These methods demonstrate that temporal context improves label efficiency and transfer. Nevertheless, reconstruction-based objectives can exploit observations on both sides of a masked timestamp, and summary-based objectives compress an interval without necessarily identifying the directed transition that maps a present state to multiple future states. SPEAR-NeXT instead asks a causal question: given only preceding latent states, what future latent trajectory is predictable?

\subsection{Predictive Latent Modeling and Multi-Horizon Objectives}

Predictive representation learning provides an alternative to reconstructing raw observations. Contrastive Predictive Coding learns representations by predicting future latent variables with an autoregressive context model \cite{oord2018representation}. I-JEPA predicts target-region embeddings from visible image context without reconstructing pixels \cite{Assran_2023_CVPR}, and V-JEPA extends feature prediction to video, showing that latent prediction can capture both appearance and motion while supporting frozen-backbone transfer \cite{Bardes_2024_TMLR}. AnySat brings a JEPA-style objective to heterogeneous EO imagery \cite{Astruc_2025_CVPR}. These works motivate prediction in representation space, but they do not directly address causal, pixel-wise forecasting of spectral Earth-state trajectories across multiple future acquisition horizons.

Observation-space Earth forecasting follows a different objective. EarthNet2021 formulates future Sentinel-2 image generation conditioned on meteorological variables as guided video prediction \cite{RequenaMesa_2021_CVPR}. Such forecasting is valuable when future reflectance or imagery is itself the desired product, but pixel-space losses must allocate capacity to radiometric detail, cloud contamination, registration variation, and other measurement-specific effects. In contrast, forecasting a spectral embedding allows the predictive objective to emphasize changes preserved by a pretrained state encoder. This does not make latent targets inherently noise-free; rather, it shifts the learning target from exact observation reproduction toward the temporal evolution represented by the encoder.

Multi-horizon forecasting is well established in time-series analysis, where architectures such as the Temporal Fusion Transformer jointly predict several future steps \cite{lim2021temporal}. In most forecasting systems, multiple horizons are output requirements and performance is measured directly at each forecast distance. SPEAR-NeXT uses multi-horizon prediction additionally as a representation constraint. Horizon-weighted latent supervision exposes the temporal model to short-range continuity, intermediate transitions, and longer-range seasonal recurrence within a single causal objective. Shorter horizons receive stronger weights for optimization stability, while longer horizons prevent the learned representation from being defined solely by adjacent-state interpolation.

Temporal attention also requires a distinction between sequence displacement and calendar phase. Rotary Position Embeddings encode relative positional interactions within self-attention \cite{Su_2024_RoFormer}; satellite time-series models additionally benefit from acquisition-date, seasonal, or phenological encodings \cite{garnot2020satellite,nyborg2022thermal,Tseng_2023_Presto}. SPEAR-NeXT therefore combines RoPE with structured month and year embeddings. RoPE describes relative progression through the observed sequence, month identifies position within the seasonal cycle, and year provides inter-annual context. The two mechanisms are complementary: observations separated by a similar temporal offset may occur in different seasonal phases, while the same calendar month may exhibit different dynamics across years.

\subsection{Positioning of SPEAR-NeXT}

SPEAR-NeXT is positioned between spectral EO foundation models, temporal satellite encoders, and joint-embedding predictive learning. Spectral foundation models learn informative representations of measurements at one or more acquisition times; temporal classifiers summarize trajectories for predefined tasks; masked temporal models recover omitted observations; and image-forecasting systems predict future measurements. SPEAR-NeXT instead decomposes pixel-wise EO pretraining into a wavelength-aware state encoder and a causal latent transition model. The temporal module predicts several future spectral states using horizon-weighted supervision and an explicit separation of relative sequence position from absolute seasonal phase.

Accordingly, the central contribution is not simply the use of a Transformer, RoPE, calendar embeddings, or multi-step prediction in isolation. It is their integration into a self-supervised objective for learning compact, predictive Earth representations: \emph{SPEAR estimates the spectral state, while SPEAR-NeXT learns how that state evolves}. This formulation is intended to produce reusable embeddings for frozen or lightweight adaptation across land-cover mapping and temporally demanding agricultural tasks, while allowing representation ablations to separate gains from temporal pretraining from those contributed by a downstream prediction head.

\section{Methodology}
\label{sec:methodology}

\subsection{Overview and Problem Formulation}
\label{sec:problem_formulation}

\begin{figure*}[!t]
    \centering
    \includegraphics[
        width=\textwidth,
        keepaspectratio
    ]{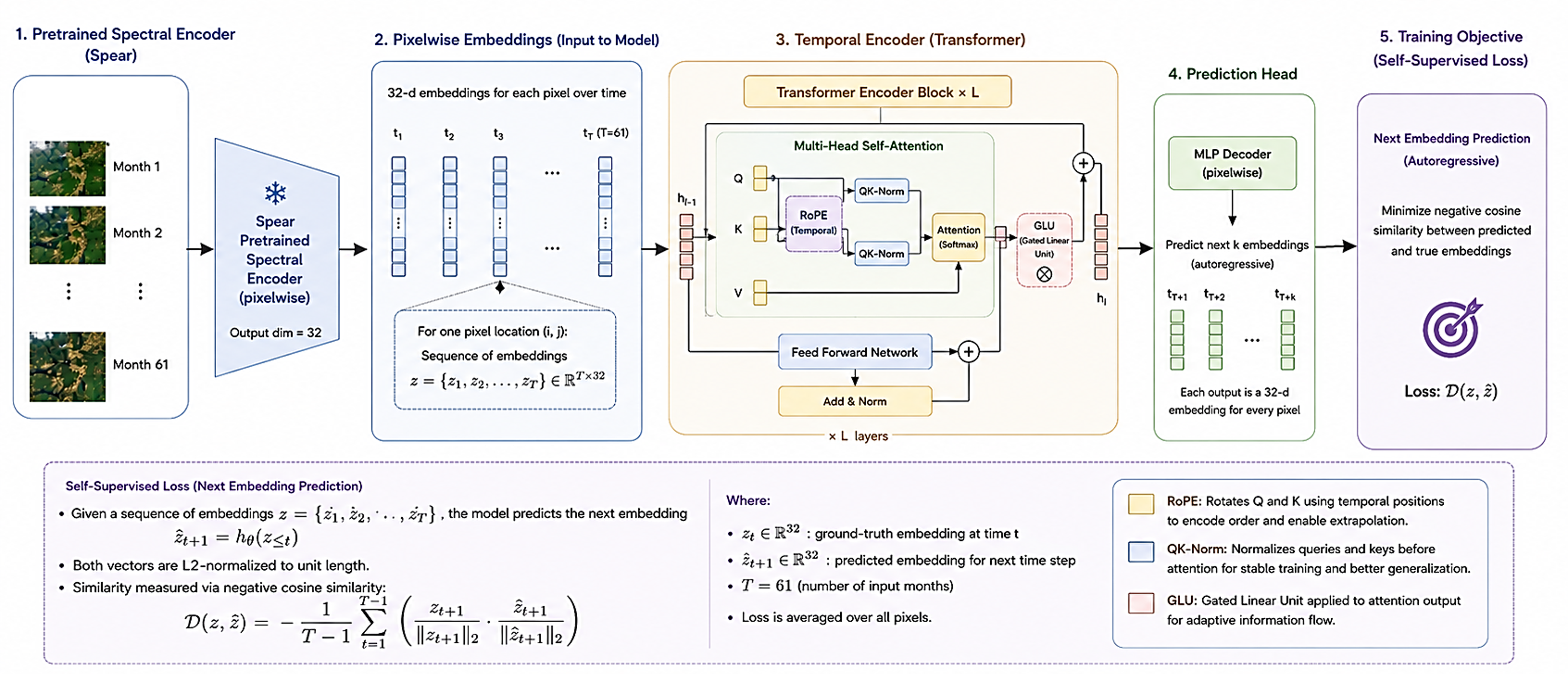}
    \caption{
    Overview of the proposed SPEAR-NeXT framework.
    A frozen pretrained SPEAR spectral encoder transforms each
    multispectral pixel observation into a compact 32-dimensional
    instantaneous spectral state. The resulting pixel-wise embedding
    sequence is processed by a causal temporal Transformer equipped
    with Rotary Position Embeddings, QK normalization, and gated
    feed-forward transformations. Horizon-specific prediction heads
    directly forecast multiple future latent states, which are optimized
    using self-supervised horizon-weighted latent trajectory supervision.
    }
    \label{fig:spear_next_framework}
\end{figure*}

Let the unlabeled pretraining corpus contain $N$ pixel-level satellite time series,
\begin{equation}
\mathcal{D}
=
\left\{
\mathcal{S}_{n}
\right\}_{n=1}^{N},
\qquad
\mathcal{S}_{n}
=
\left\{
\left(
\mathbf{x}_{n,t},
\tau_{n,t}
\right)
\right\}_{t=1}^{T_n},
\label{eq:dataset}
\end{equation}
where $\mathbf{x}_{n,t}\in\mathbb{R}^{C}$ is the multispectral observation of pixel $n$ at timestamp $\tau_{n,t}$, $C$ is the number of spectral bands, and $T_n$ is the number of valid observations available for that pixel. Observations are chronologically ordered such that
$\tau_{n,1}<\cdots<\tau_{n,T_n}$. The formulation permits different sequence lengths across pixels.

SPEAR-NeXT decomposes representation learning into two complementary stages. First, a pretrained wavelength-aware SPEAR encoder estimates an instantaneous spectral state from each multispectral observation. Second, a causal temporal model learns how sequences of these states evolve and predicts their future latent trajectory. For every horizon $k\in\{1,\ldots,K\}$, the objective is to estimate the state at time $t+k$ using only observations available up to time $t$:
\begin{equation}
\widehat{\mathbf{e}}_{n,t+k\mid t}
=
g_{\omega_k}
\left(
q_{\theta}
\left(
\mathbf{e}_{n,1:t},
\boldsymbol{\tau}_{n,1:t}
\right)
\right),
\label{eq:overall_prediction}
\end{equation}
where $\mathbf{e}_{n,t}\in\mathbb{R}^{D}$ is the spectral state, $q_{\theta}$ is the causal temporal encoder, and $g_{\omega_k}$ is the predictor associated with horizon $k$. The notation $t+k\mid t$ explicitly distinguishes the predicted future state from the causal context used to estimate it.

Throughout this work, \emph{causal} refers to autoregressive information flow enforced by a temporal attention mask. It does not imply intervention-based causal inference. Similarly, \emph{Earth state} denotes a learned compact spectral representation rather than a complete physical description of the land surface.

\subsection{Instantaneous Spectral States from SPEAR}
\label{sec:spear_states}

The instantaneous spectral encoder is adopted from the previously published SPEAR framework \cite{Ranjan_2026_WACV}. SPEAR introduced a wavelength-conditioned Spectral-MAE that operates directly on per-pixel multispectral vectors and uses continuous band-center wavelength and full width at half maximum (FWHM) metadata during tokenization. Its masked-band pretraining procedure, Fourier spectral metadata encoding, encoder--decoder architecture, and reconstruction objective are described in \cite{Ranjan_2026_WACV} and are therefore not repeated here.

Although the original SPEAR framework supports optical, radar, climate, and multimodal fusion, the temporal formulation studied in this work uses its pretrained optical Spectral-MAE branch. Let
\begin{equation}
\boldsymbol{\lambda}
=
\left[
\lambda_1,\ldots,\lambda_C
\right]^{\top},
\qquad
\boldsymbol{\Delta\lambda}
=
\left[
\Delta\lambda_1,\ldots,\Delta\lambda_C
\right]^{\top}
\end{equation}
denote the band-center wavelengths and FWHM values. Each observation is mapped to a compact state through
\begin{equation}
\mathbf{e}_{n,t}
=
f_{\psi}^{\mathrm{SPEAR}}
\left(
\mathbf{x}_{n,t};
\boldsymbol{\lambda},
\boldsymbol{\Delta\lambda}
\right),
\qquad
\mathbf{e}_{n,t}\in\mathbb{R}^{D}.
\label{eq:spear_embedding}
\end{equation}
In the present implementation, $D=32$, consistent with the compact SPEAR spectral descriptor. The reconstruction decoder used during SPEAR pretraining is discarded, and the unmasked encoder output associated with the spectral class token is used as $\mathbf{e}_{n,t}$.

The pretrained parameters $\psi$ are held fixed during temporal pretraining. Consequently, the spectral states can be computed and cached once, reducing temporal training cost and ensuring that future-state targets remain stationary.

\subsection{Temporal Token Construction}
\label{sec:temporal_tokens}

Satellite time series contain two complementary forms of temporal information. Relative temporal progression describes ordering and displacement between observations, whereas absolute calendar phase describes seasonal position and inter-annual context. SPEAR-NeXT models these signals separately.

Let $m(\tau_{n,t})\in\{1,\ldots,12\}$ and $y(\tau_{n,t})\in\{1,\ldots,N_y\}$ denote the month and indexed year associated with timestamp $\tau_{n,t}$. Learnable month and year embeddings are defined as
\begin{equation}
\mathbf{m}_{n,t}
=
\mathbf{M}_{m(\tau_{n,t})},
\qquad
\mathbf{y}_{n,t}
=
\mathbf{Y}_{y(\tau_{n,t})},
\label{eq:calendar_embeddings}
\end{equation}
where
$\mathbf{M}\in\mathbb{R}^{12\times D_m}$ and
$\mathbf{Y}\in\mathbb{R}^{N_y\times D_y}$.
The absolute calendar representation is
\begin{equation}
\mathbf{c}_{n,t}
=
\left[
\mathbf{m}_{n,t};
\mathbf{y}_{n,t}
\right]
\in
\mathbb{R}^{D_m+D_y}.
\end{equation}
The spectral state and calendar representation are concatenated and projected into the temporal model dimension:
\begin{equation}
\mathbf{h}_{n,t}^{(0)}
=
\mathbf{W}_{\mathrm{in}}
\left[
\mathbf{e}_{n,t};
\mathbf{c}_{n,t}
\right]
+
\mathbf{b}_{\mathrm{in}},
\qquad
\mathbf{h}_{n,t}^{(0)}
\in
\mathbb{R}^{D_T}.
\label{eq:temporal_input}
\end{equation}
Month and year embeddings therefore provide absolute temporal context, while relative progression is incorporated within self-attention through Rotary Position Embeddings.

\subsection{Causal Temporal Dynamics Encoder}
\label{sec:temporal_encoder}

Let
\begin{equation}
\mathbf{H}_{n}^{(\ell)}
=
\left[
\mathbf{h}_{n,1}^{(\ell)},
\ldots,
\mathbf{h}_{n,T_n}^{(\ell)}
\right]^{\top}
\in
\mathbb{R}^{T_n\times D_T}
\end{equation}
denote the sequence at layer $\ell$. SPEAR-NeXT employs a pre-normalized causal Transformer with $L$ layers and $A$ attention heads.

\subsubsection{QK-normalized rotary self-attention}

For attention head $a$, the query, key, and value projections are
\begin{align}
\mathbf{Q}_{a}^{(\ell)}
&=
\operatorname{QKNorm}
\left(
\operatorname{LN}
\left(
\mathbf{H}_{n}^{(\ell)}
\right)
\mathbf{W}_{Q,a}^{(\ell)}
\right),
\\
\mathbf{K}_{a}^{(\ell)}
&=
\operatorname{QKNorm}
\left(
\operatorname{LN}
\left(
\mathbf{H}_{n}^{(\ell)}
\right)
\mathbf{W}_{K,a}^{(\ell)}
\right),
\\
\mathbf{V}_{a}^{(\ell)}
&=
\operatorname{LN}
\left(
\mathbf{H}_{n}^{(\ell)}
\right)
\mathbf{W}_{V,a}^{(\ell)}.
\label{eq:qkv}
\end{align}
QK normalization is applied independently to each query and key head vector:
\begin{equation}
\operatorname{QKNorm}
\left(
\mathbf{a}
\right)
=
\boldsymbol{\gamma}
\odot
\frac{
\mathbf{a}-\mu(\mathbf{a})
}{
\sqrt{
\sigma^2(\mathbf{a})+\varepsilon
}
},
\label{eq:qk_norm}
\end{equation}
where $\mu(\cdot)$ and $\sigma^2(\cdot)$ are computed across the head dimension, $\boldsymbol{\gamma}$ is a learnable scale, and $\varepsilon$ is a numerical-stability constant. Layer normalization controls the scale of token activations before projection, whereas QK normalization directly controls the query--key dot products used by attention.

Let $p_{n,t}$ denote the temporal coordinate of observation $t$. For regularly composited sequences, $p_{n,t}=t$; for irregular acquisitions, it may be defined using elapsed time from the beginning of the sequence. RoPE applies a position-dependent rotation to the query and key vectors:
\begin{equation}
\widetilde{\mathbf{q}}_{n,t,a}^{(\ell)}
=
\mathbf{R}
\left(
p_{n,t}
\right)
\mathbf{q}_{n,t,a}^{(\ell)},
\qquad
\widetilde{\mathbf{k}}_{n,t,a}^{(\ell)}
=
\mathbf{R}
\left(
p_{n,t}
\right)
\mathbf{k}_{n,t,a}^{(\ell)}.
\label{eq:rope}
\end{equation}
The identity
\begin{equation}
\mathbf{R}(p_i)^{\top}\mathbf{R}(p_j)
=
\mathbf{R}(p_j-p_i)
\end{equation}
allows the attention interaction to depend on relative temporal displacement.

Causality is enforced through the mask
\begin{equation}
\mathcal{C}_{ij}
=
\begin{cases}
0, & j\leq i,\\
-\infty, & j>i.
\end{cases}
\label{eq:causal_mask}
\end{equation}
An additional mask $\mathcal{P}_{ij}$ excludes padded or invalid timestamps when variable-length sequences are batched. The output of head $a$ is
\begin{equation}
\mathbf{A}_{a}^{(\ell)}
=
\operatorname{softmax}
\left(
\frac{
\widetilde{\mathbf{Q}}_{a}^{(\ell)}
\widetilde{\mathbf{K}}_{a}^{(\ell)\top}
}{
\sqrt{d_h}
}
+
\boldsymbol{\mathcal{C}}
+
\boldsymbol{\mathcal{P}}
\right)
\mathbf{V}_{a}^{(\ell)},
\label{eq:causal_attention}
\end{equation}
where $d_h=D_T/A$ is the dimension of each attention head. Hence, the representation at time $t$ is computed only from observations at times $1,\ldots,t$.

The multi-head attention output is combined through a residual connection:
\begin{equation}
\mathbf{H}_{n}^{(\ell+\frac{1}{2})}
=
\mathbf{H}_{n}^{(\ell)}
+
\operatorname{Concat}
\left(
\mathbf{A}_{1}^{(\ell)},
\ldots,
\mathbf{A}_{A}^{(\ell)}
\right)
\mathbf{W}_{O}^{(\ell)}.
\label{eq:attention_residual}
\end{equation}

\subsubsection{SwiGLU feed-forward transformation}

The feed-forward sublayer uses a gated SwiGLU transformation:
\begin{equation}
\operatorname{SwiGLU}
\left(
\mathbf{a}
\right)
=
\mathbf{W}_{2}
\left[
\operatorname{SiLU}
\left(
\mathbf{W}_{g}\mathbf{a}
+
\mathbf{b}_{g}
\right)
\odot
\left(
\mathbf{W}_{u}\mathbf{a}
+
\mathbf{b}_{u}
\right)
\right]
+
\mathbf{b}_{2},
\label{eq:swiglu}
\end{equation}
where $\odot$ denotes element-wise multiplication. The layer output is
\begin{equation}
\mathbf{H}_{n}^{(\ell+1)}
=
\mathbf{H}_{n}^{(\ell+\frac{1}{2})}
+
\operatorname{SwiGLU}
\left(
\operatorname{LN}
\left(
\mathbf{H}_{n}^{(\ell+\frac{1}{2})}
\right)
\right).
\label{eq:ffn_residual}
\end{equation}
After $L$ Transformer layers, the contextual state at time $t$ is
\begin{equation}
\mathbf{r}_{n,t}
=
\operatorname{LN}
\left(
\mathbf{h}_{n,t}^{(L)}
\right)
\in
\mathbb{R}^{D_T}.
\label{eq:context_state}
\end{equation}

\subsection{Causal Multi-Horizon Latent Forecasting}
\label{sec:multi_horizon}

A separate lightweight prediction head is used for each horizon. For $k\in\{1,\ldots,K\}$,
\begin{equation}
g_{\omega_k}
\left(
\mathbf{r}
\right)
=
\mathbf{W}_{k,2}
\operatorname{SiLU}
\left(
\mathbf{W}_{k,1}\mathbf{r}
+
\mathbf{b}_{k,1}
\right)
+
\mathbf{b}_{k,2},
\label{eq:horizon_head}
\end{equation}
and the future-state prediction is
\begin{equation}
\widehat{\mathbf{e}}_{n,t+k\mid t}
=
g_{\omega_k}
\left(
\mathbf{r}_{n,t}
\right).
\label{eq:horizon_prediction}
\end{equation}
The horizon $k$ denotes a future sequence step; under a fixed monthly compositing scheme, it corresponds to $k$ months. Horizon-specific heads permit the mapping from contextual state to target state to vary with forecast distance and avoid recursive error accumulation.

The corresponding target is generated by the frozen SPEAR encoder:
\begin{equation}
\mathbf{e}_{n,t+k}^{\star}
=
\operatorname{sg}
\left[
f_{\psi}^{\mathrm{SPEAR}}
\left(
\mathbf{x}_{n,t+k};
\boldsymbol{\lambda},
\boldsymbol{\Delta\lambda}
\right)
\right],
\label{eq:target_state}
\end{equation}
where $\operatorname{sg}[\cdot]$ denotes stop-gradient. Because $f_{\psi}^{\mathrm{SPEAR}}$ is frozen, the target space remains fixed throughout temporal optimization.

\subsection{Horizon-Weighted Latent Trajectory Objective}
\label{sec:trajectory_loss}

For each horizon $k$, define the set of valid context--target pairs as
\begin{equation}
\Omega_k
=
\left\{
(n,t)
\;\middle|\;
1\leq n\leq N,\;
1\leq t\leq T_n-k
\right\}.
\label{eq:valid_pairs}
\end{equation}
Directional agreement is optimized using cosine distance:
\begin{equation}
\begin{aligned}
\mathcal{L}_{\mathrm{cos}}^{(k)}
&=
\frac{1}{|\Omega_k|}
\sum_{(n,t)\in\Omega_k}
\\[-0.5ex]
&\quad\left[
1-
\frac{
\widehat{\mathbf{e}}_{n,t+k\mid t}^{\top}
\mathbf{e}_{n,t+k}^{\star}
}{
\bigl(\lVert\widehat{\mathbf{e}}_{n,t+k\mid t}\rVert_2+\varepsilon\bigr)
\bigl(\lVert\mathbf{e}_{n,t+k}^{\star}\rVert_2+\varepsilon\bigr)
}
\right].
\end{aligned}
\label{eq:cosine_loss}
\end{equation}
Since cosine distance is insensitive to vector scale, it is complemented by latent vector regression:
\begin{equation}
\mathcal{L}_{\mathrm{reg}}^{(k)}
=
\frac{1}{D|\Omega_k|}
\sum_{(n,t)\in\Omega_k}
\left\|
\widehat{\mathbf{e}}_{n,t+k\mid t}
-
\mathbf{e}_{n,t+k}^{\star}
\right\|_2^2.
\label{eq:regression_loss}
\end{equation}
The regression term penalizes deviations in both direction and scale and therefore limits embedding-magnitude drift that is not constrained by cosine alignment alone. The loss at horizon $k$ is
\begin{equation}
\mathcal{L}^{(k)}
=
\alpha
\mathcal{L}_{\mathrm{cos}}^{(k)}
+
(1-\alpha)
\mathcal{L}_{\mathrm{reg}}^{(k)},
\qquad
0\leq\alpha\leq1.
\label{eq:horizon_loss}
\end{equation}

The contribution of each horizon is controlled by
\begin{equation}
w_k
=
\frac{k^{-\gamma}}
{\sum_{j=1}^{K}j^{-\gamma}},
\qquad
\gamma\geq0.
\label{eq:horizon_weight}
\end{equation}
The inverse-horizon weighting used in this work is obtained with $\gamma=1$, while $\gamma=0$ yields uniform weighting. The complete temporal pretraining objective is
\begin{equation}
\mathcal{L}_{\mathrm{temp}}
=
\sum_{k=1}^{K}
w_k
\mathcal{L}^{(k)}.
\label{eq:temporal_loss}
\end{equation}
Shorter horizons receive stronger supervision for optimization stability, whereas intermediate and longer horizons require the contextual state to retain information that remains predictive beyond adjacent observations. Multi-horizon forecasting is therefore used as a representation-learning constraint in addition to being a prediction objective.

\subsection{Optimization and Transfer}
\label{sec:optimization_transfer}

SPEAR-NeXT is trained after completion of SPEAR spectral pretraining. The spectral encoder is fixed and only the temporal encoder and prediction heads are optimized:
\begin{equation}
\left(
\theta^{\star},
\omega_{1:K}^{\star}
\right)
=
\arg\min_{\theta,\omega_{1:K}}
\mathcal{L}_{\mathrm{temp}}
\quad
\text{subject to}
\quad
\psi=\psi^{\star}.
\label{eq:optimization}
\end{equation}
This modular training procedure separates instantaneous spectral state estimation from temporal state evolution, enables reusable cached embeddings, and avoids instability caused by a moving target representation.

The horizon-specific prediction heads are used only during self-supervised temporal pretraining. For downstream transfer, the contextual representation $\mathbf{r}_{n,t}$ is retained. At a forecasting cutoff $t$, it is a strictly causal representation derived only from observations available up to that time. For tasks using a complete sequence, the final valid contextual state or a task-specific aggregation over
$\{\mathbf{r}_{n,t}\}_{t=1}^{T_n}$ is supplied to a lightweight downstream model. The SPEAR and temporal encoders may be frozen for representation probing or fine-tuned under a task-specific adaptation protocol.

\section{Experimental Analysis}
\label{sec:experimental_analysis}

This section evaluates whether SPEAR-NeXT learns temporally predictive and geometrically stable pixel-level representations. The analysis is organized around four questions: (i) how forecasting quality changes with prediction horizon, (ii) whether future-state predictions preserve embedding direction and magnitude, (iii) whether the learned dynamics retain seasonal temporal geometry, and (iv) whether gains in downstream transfer arise from temporal pretraining rather than from additional prediction-head capacity. Unless otherwise stated, all forecasting metrics are computed on held-out pixel locations that are not used during model optimization.

\subsection{Dataset Construction and Preprocessing}
\label{sec:dataset_construction}

Two geographically independent pretraining and evaluation pipelines are
constructed for India and the contiguous United States (CONUS). The two
regional corpora are not pooled during pretraining. Instead, an independent
SPEAR-NeXT model is pretrained for each region and subsequently evaluated
only on the downstream datasets associated with that region. This design
allows the temporal representations to be assessed under distinct
geographic, climatic, agricultural, and land-cover distributions.

The India corpus contains approximately $2.7$ million pixel-level samples,
whereas the CONUS corpus contains approximately $4.5$ million samples.
Both corpora cover the period from January 2020 to December 2024 and contain
$T=60$ monthly observations per retained pixel location:
\begin{equation}
T = 12 \times 5 = 60.
\end{equation}
The complete composition of the two corpora and their associated downstream
datasets is summarized in Table~\ref{tab:regional_dataset_summary}.
Additional dataset descriptions, geographic visualizations, and
source-specific details are provided in
Appendix~\ref{app:dataset_information}.

\begin{table*}[t]
\centering
\caption{
Summary of the two independent regional pretraining and evaluation
pipelines. Each region is used to pretrain an independent SPEAR-NeXT model,
and no samples or model parameters are shared across regions.
}
\label{tab:regional_dataset_summary}
\resizebox{\textwidth}{!}{
\begin{tabular}{llllllll}
\toprule
Region &
Model &
Pretraining size &
Temporal span &
Pretraining modalities &
Representation &
Downstream datasets \\
\midrule

India &
SPEAR-NeXT$_{\text{India}}$ &
$\sim$2.7M pixels &
2020--2024
(60 months) &
Sentinel-2 \cite{Drusch_2012_Sentinel2},
Sentinel-1 \cite{Torres_2012_Sentinel1},
ERA5-Land \cite{MunozSabater_2021_ERA5Land} &
$60\times32$ fused pixel states &
SICKLE \cite{Sani_2024_WACV},
Sen1Floods11 \cite{Bonafilia_2020_CVPRW},
VIIRS \cite{Schroeder_2014_VIIRS},
Dynamic World \cite{Brown_2022_DynamicWorld}
\\

CONUS &
SPEAR-NeXT$_{\text{CONUS}}$ &
$\sim$4.5M pixels &
2020--2024
(60 months) &
Sentinel-2 \cite{Drusch_2012_Sentinel2},
Sentinel-1 \cite{Torres_2012_Sentinel1},
ERA5-Land \cite{MunozSabater_2021_ERA5Land} &
$60\times32$ fused pixel states &
USDA NASS \cite{USDA_NASS_QuickStats},
CropHarvest \cite{Tseng_2021_CropHarvest},
Sen1Floods11 \cite{Bonafilia_2020_CVPRW},
Dynamic World \cite{Brown_2022_DynamicWorld}
\\

\bottomrule
\end{tabular}}
\end{table*}

\paragraph{Monthly multimodal observations.}

Let
\begin{equation}
r\in\mathcal{R}
=
\left\{
\mathrm{India},
\mathrm{CONUS}
\right\}
\end{equation}
denote a regional corpus. For pixel location $n$, the monthly multimodal
sequence is defined as
\begin{equation}
\mathcal{S}_{n}^{(r)}
=
\left\{
\left(
\mathbf{x}_{n,t}^{\mathrm{S2},(r)},
\mathbf{x}_{n,t}^{\mathrm{S1},(r)},
\mathbf{x}_{n,t}^{\mathrm{ERA5},(r)},
\tau_t
\right)
\right\}_{t=1}^{60}.
\label{eq:regional_multimodal_sequence}
\end{equation}

Here,
\begin{equation}
\mathbf{x}_{n,t}^{\mathrm{S2},(r)}
\in\mathbb{R}^{10}
\end{equation}
contains the ten selected Sentinel-2 multispectral bands
\cite{Drusch_2012_Sentinel2},
\begin{equation}
\mathbf{x}_{n,t}^{\mathrm{S1},(r)}
\in\mathbb{R}^{2}
\end{equation}
contains Sentinel-1 VV and VH radar backscatter
\cite{Torres_2012_Sentinel1}, and
\begin{equation}
\mathbf{x}_{n,t}^{\mathrm{ERA5},(r)}
\in\mathbb{R}^{C_{\mathrm{clim}}}
\end{equation}
contains the matched ERA5/ERA5-Land climate and environmental variables
\cite{MunozSabater_2021_ERA5Land}. These include the temperature,
precipitation, and elevation-related variables used in the SPEAR
multimodal representation pipeline \cite{Ranjan_2026_WACV}. The
observations from all three sources are aligned to a common monthly temporal
grid.

The original India SPEAR corpus additionally contains PlanetScope SuperDove
observations used during multisensor spectral pretraining
\cite{Planet_2026_PlanetScope,Ranjan_2026_WACV}. PlanetScope is not assumed
to be available at every timestep of the uniform 2020--2024 monthly
sequence and is therefore not included as a required input to the
SPEAR-NeXT temporal model.

\paragraph{Regional pixel-state construction.}

At every timestep, the three monthly modalities are processed using their
corresponding pretrained SPEAR components. The optical representation is
obtained using the wavelength-aware Spectral-MAE:
\begin{equation}
\mathbf{e}_{n,t}^{\mathrm{S2},(r)}
=
f_{\psi_{\mathrm{S2}}}^{(r)}
\left(
\mathbf{x}_{n,t}^{\mathrm{S2},(r)};
\boldsymbol{\lambda},
\boldsymbol{\Delta\lambda}
\right),
\end{equation}
where $\boldsymbol{\lambda}$ and
$\boldsymbol{\Delta\lambda}$ denote the Sentinel-2 band-center wavelengths
and bandwidths. The radar and environmental representations are
\begin{align}
\mathbf{e}_{n,t}^{\mathrm{S1},(r)}
&=
f_{\psi_{\mathrm{S1}}}^{(r)}
\left(
\mathbf{x}_{n,t}^{\mathrm{S1},(r)}
\right),
\\
\mathbf{e}_{n,t}^{\mathrm{ERA5},(r)}
&=
f_{\psi_{\mathrm{ERA5}}}^{(r)}
\left(
\mathbf{x}_{n,t}^{\mathrm{ERA5},(r)},
\mathbf{g}_{n}
\right),
\end{align}
where $\mathbf{g}_{n}$ denotes the geographic metadata associated with
pixel $n$. These representations are produced by the Spectral-MAE,
BYOL-Denoise, and Climate-MAE components of SPEAR, respectively
\cite{Ranjan_2026_WACV}.

The modality-specific representations are combined using the regional
SPEAR fusion operator:
\begin{equation}
\mathbf{e}_{n,t}^{(r)}
=
\mathcal{F}_{\phi}^{(r)}
\left(
\mathbf{e}_{n,t}^{\mathrm{S2},(r)},
\mathbf{e}_{n,t}^{\mathrm{S1},(r)},
\mathbf{e}_{n,t}^{\mathrm{ERA5},(r)}
\right),
\qquad
\mathbf{e}_{n,t}^{(r)}
\in\mathbb{R}^{32}.
\label{eq:regional_fused_state}
\end{equation}
The resulting temporal input to SPEAR-NeXT is
\begin{equation}
\mathbf{E}_{n}^{(r)}
=
\left[
\mathbf{e}_{n,1}^{(r)},
\mathbf{e}_{n,2}^{(r)},
\ldots,
\mathbf{e}_{n,60}^{(r)}
\right]^{\top}
\in
\mathbb{R}^{60\times32}.
\label{eq:regional_embedding_sequence}
\end{equation}
The modality encoders and fusion module are frozen during temporal
pretraining, and the extracted monthly pixel states are cached before
training the corresponding regional temporal model.

\paragraph{Preprocessing and quality control.}

Only pixel locations satisfying the required temporal-coverage and
data-quality criteria are retained. Invalid and cloud-contaminated
Sentinel-2 observations are removed during optical preprocessing.
Sentinel-1 VV and VH observations are processed consistently in the
calibrated backscatter domain, while ERA5/ERA5-Land variables are matched
to the same pixel location and monthly temporal index.

A separate preprocessing transformation is fitted for each modality and
region. For modality
\begin{equation}
m
\in
\left\{
\mathrm{S2},
\mathrm{S1},
\mathrm{ERA5}
\right\},
\end{equation}
the standardized observation is
\begin{equation}
\widetilde{\mathbf{x}}_{n,t}^{m,(r)}
=
\mathcal{T}_{m}^{(r)}
\left(
\mathbf{x}_{n,t}^{m,(r)}
\right),
\end{equation}
where the RobustScaler transformation
$\mathcal{T}_{m}^{(r)}$ is fitted exclusively on the corresponding
regional training partition. The fitted transformation is then applied
unchanged to the validation and test partitions. No normalization
statistics are shared between India and CONUS, and future observations are
never used to normalize or impute earlier timestamps.

\paragraph{Leakage-controlled regional partitioning.}

Each regional corpus is independently divided into training, validation,
and test sets using a $60{:}20{:}20$ split:
\begin{equation}
\mathcal{D}^{(r)}
=
\mathcal{D}_{\mathrm{train}}^{(r)}
\cup
\mathcal{D}_{\mathrm{val}}^{(r)}
\cup
\mathcal{D}_{\mathrm{test}}^{(r)},
\end{equation}
with
\begin{equation}
\left|
\mathcal{D}_{\mathrm{train}}^{(r)}
\right|
:
\left|
\mathcal{D}_{\mathrm{val}}^{(r)}
\right|
:
\left|
\mathcal{D}_{\mathrm{test}}^{(r)}
\right|
=
60:20:20.
\end{equation}

Partitioning is performed at the pixel-location level rather than at the
individual pixel--month level. Therefore, all $60$ monthly observations
belonging to one pixel are assigned to the same partition. Test pixels are
kept geographically separate from training pixels to prevent spatial
leakage from neighbouring locations with highly correlated spectral,
radar, climate, and temporal trajectories.

Temporal leakage is prevented by ensuring that no observation belonging to
a held-out test sequence contributes to model optimization, target
construction, normalization, model selection, or early stopping. When a
downstream experiment includes an explicit temporal holdout, the
corresponding years or monthly periods are also excluded from the training
partition. Thus, the reported test results are independent at both the
pixel-location and temporal-evaluation levels.

\paragraph{Independent regional models and evaluations.}

The India and CONUS temporal models are optimized independently:
\begin{align}
\left(
\theta_{\mathrm{India}}^{\star},
\omega_{\mathrm{India},1:K}^{\star}
\right)
&=
\arg\min_{\theta,\omega_{1:K}}
\mathcal{L}_{\mathrm{temp}}^{\mathrm{India}},
\\
\left(
\theta_{\mathrm{CONUS}}^{\star},
\omega_{\mathrm{CONUS},1:K}^{\star}
\right)
&=
\arg\min_{\theta,\omega_{1:K}}
\mathcal{L}_{\mathrm{temp}}^{\mathrm{CONUS}}.
\end{align}
No temporal-model parameters, pretraining samples, validation samples, or
test samples are shared between the two regional pipelines.

The India-pretrained model is evaluated on SICKLE crop classification
\cite{Sani_2024_WACV}, Sen1Floods11 flood detection
\cite{Bonafilia_2020_CVPRW}, VIIRS-based fire-related classification
\cite{Schroeder_2014_VIIRS}, and Dynamic World land-cover classification
\cite{Brown_2022_DynamicWorld}. The CONUS-pretrained model is evaluated
separately on USDA NASS crop-yield prediction
\cite{USDA_NASS_QuickStats}, CropHarvest crop/non-crop classification
\cite{Tseng_2021_CropHarvest}, Sen1Floods11 flood detection
\cite{Bonafilia_2020_CVPRW}, and Dynamic World land-cover classification
\cite{Brown_2022_DynamicWorld}.

Model selection and early stopping use only the validation partition of the
corresponding region. Forecasting, representation, and downstream
performance are reported using that region's held-out test data. Detailed
descriptions and geographic visualizations of the pretraining and
downstream datasets are provided in
Appendix~\ref{app:dataset_information}.

\subsection{Training Configuration}
\label{sec:training_configuration}

The temporal encoder contains $L=4$ causal Transformer layers with temporal model dimension $D_T=128$ and $A=4$ self-attention heads. Direct prediction is performed for $K=46$ future sequence steps using horizon-specific prediction heads. The SPEAR encoder remains frozen throughout temporal pretraining, ensuring that improvements can be attributed to temporal representation learning and that the target embedding space does not drift during optimization.

The temporal encoder and prediction heads are optimized using AdamW with linear learning-rate warmup, gradient clipping, and mixed-precision training. An exponential moving average (EMA) copy of the temporal parameters is maintained and used for validation and final evaluation. The composite objective in Eq.~\eqref{eq:temporal_loss} uses $\alpha=0.5$, assigning equal weight to cosine alignment and latent vector regression. Unless otherwise stated, inverse-horizon weighting is used with $\gamma=1$:
\begin{equation}
w_k
=
\frac{k^{-1}}
{\sum_{j=1}^{K}j^{-1}}.
\end{equation}
The exact learning rate, weight decay, batch size, warmup duration, gradient-clipping threshold, EMA decay, number of epochs, and early-stopping patience are reported in Table~\ref{tab:training_hyperparameters}.

\begin{table}[t]
\centering
\footnotesize
\setlength{\tabcolsep}{4pt}
\renewcommand{\arraystretch}{1.08}

\caption{
Temporal pretraining configuration of the SPEAR-NeXT temporal module.
}
\label{tab:training_hyperparameters}

\begin{tabularx}{\columnwidth}{
    @{}
    >{\raggedright\arraybackslash}p{0.48\columnwidth}
    Y
    @{}
}
\toprule
\textbf{Configuration} & \textbf{Value} \\
\midrule

Spectral embedding dimension $D$
    & 32 \\

Temporal model dimension $D_T$
    & 128 \\

Transformer layers $L$
    & 4 \\

Attention heads $A$
    & 4 \\

Maximum prediction horizon $K$
    & 46 \\

Loss balance $\alpha$
    & 0.5 \\

Horizon exponent $\gamma$
    & 1 \\

Optimizer
    & AdamW
      $(\beta_1=0.9,\ \beta_2=0.95)$ \\

Base learning rate
    & $1\times10^{-4}$ \\

Weight decay
    & 0.05 \\

Batch size
    & 256 \\

Warmup epochs
    & 5 \\

Gradient-clipping threshold
    & 1.0 \\

EMA decay
    & 0.999 \\

Training epochs
    & 20 \\

Early-stopping patience
    & 5
      $(\mathrm{min}\text{-}\Delta=1\times10^{-5})$ \\

\bottomrule
\end{tabularx}
\end{table}

\begin{table}[t]
\centering
\footnotesize
\setlength{\tabcolsep}{4pt}
\renewcommand{\arraystretch}{1.08}

\caption{
Architectural details of the SPEAR-NeXT multi-horizon temporal predictor
built on frozen SPEAR representations~\cite{Ranjan_2026_WACV}.
}
\label{tab:architecture_details}

\begin{tabularx}{\columnwidth}{
    @{}
    >{\raggedright\arraybackslash}p{0.38\columnwidth}
    Y
    @{}
}
\toprule
\textbf{Component} & \textbf{Specification} \\
\midrule

Backbone
    & Frozen SPEAR encoder
      $(D=32)$ \\

Temporal encoding
    & Learned month and year embeddings
      $(16+16\text{-d})$ \\

Positional encoding
    & RoPE $(\theta=10{,}000)$ \\

Attention stabilization
    & QK-Norm \\

Attention masking
    & Causal \\

Feed-forward network
    & SwiGLU with hidden dimension 512 \\

Encoder layers $L$
    & 4 \\

Attention heads $A$
    & 4
      $(\text{head dimension}=32)$ \\

Model dimension $D_T$
    & 128 \\

Dropout
    & 0.1 \\

Prediction horizons $K$
    & 46 independent prediction heads \\

Horizon weighting
    & $w_k\propto 1/k$ \\

Loss function
    & $\alpha\mathcal{L}_{\mathrm{cos}}
      +(1-\alpha)\mathcal{L}_{\mathrm{MSE}}$,
      computed independently at each horizon \\

Target handling
    & Stop-gradient target embeddings
      (BYOL-style) \\

\bottomrule
\end{tabularx}
\end{table}

The temporal predictor sits atop a frozen SPEAR spectral encoder and treats each pixel's monthly embedding sequence as input to a causal transformer. Table~\ref{tab:architecture_details} summarizes the full architectural configuration. We adopt a small set of stabilization and efficiency techniques originally developed for large language models, RoPE for relative positional encoding, QK-Norm for attention stability, and SwiGLU feed-forward layers, motivated by the fact that satellite pixel time series, unlike text corpora, offer a comparatively data-scarce training regime where such stability tricks matter more, not less. The core contribution is Multi-Horizon Prediction (MHP): rather than predicting only the next-step embedding, the model attaches $K{=}46$ independent lightweight MLP heads to the shared transformer trunk, one per horizon $k=1,\dots,46$, each trained to forecast the SPEAR embedding at $t+k$ from context up to $t$. Horizons are weighted inversely ($w_k \propto 1/k$) so that near-term, higher-confidence predictions dominate the loss while long-range horizons still receive gradient signal. Each horizon's loss combines a cosine (directional) term and an MSE (magnitude) term, balanced by $\alpha$. Targets are stop-gradiented from the same frozen SPEAR encoder. This is BYOL-inspired in spirit (no explicit target reconstruction, prediction in representation space) but does not use BYOL's dual online/momentum-teacher architecture; the EMA copy of the model maintained during training is used only for stable validation and inference, not as a separate target network. Because the SPEAR backbone remains fully frozen throughout this pretraining stage, all representational adaptation for temporal dynamics happens exclusively within the lightweight NEPA transformer head, keeping the additional training cost modest relative to the frozen encoder.

\subsection{Forecasting Baselines and Ablation Protocol}
\label{sec:forecasting_baselines}

Embedding forecasts are compared against simple temporal baselines to determine whether the learned model improves upon continuity and seasonality priors.

\paragraph{Persistence.}
The most recent observed state is copied to every future horizon:
\begin{equation}
\widehat{\mathbf{e}}_{n,t+k\mid t}^{\mathrm{pers}}
=
\mathbf{e}_{n,t}.
\label{eq:persistence_baseline}
\end{equation}

\paragraph{Linear latent extrapolation.}
The most recent latent displacement is extended to horizon $k$:
\begin{equation}
\widehat{\mathbf{e}}_{n,t+k\mid t}^{\mathrm{lin}}
=
\mathbf{e}_{n,t}
+
k
\left(
\mathbf{e}_{n,t}
-
\mathbf{e}_{n,t-1}
\right),
\qquad
t\geq2.
\label{eq:linear_baseline}
\end{equation}

\paragraph{Seasonal persistence.}
When the sequence cadence admits a known seasonal period $P$ and the corresponding state lies within the observed context, the state from the previous cycle is used:
\begin{equation}
\widehat{\mathbf{e}}_{n,t+k\mid t}^{\mathrm{season}}
=
\mathbf{e}_{n,t+k-P},
\qquad
t+k-P\leq t.
\label{eq:seasonal_baseline}
\end{equation}
For monthly composites, $P=12$.

The temporal objective is further examined using controlled ablations: cosine-only versus regression-only versus composite loss; uniform versus inverse-horizon weighting; single-step versus multi-horizon prediction; RoPE-only, calendar-only, and combined temporal conditioning; and models trained with or without QK normalization. All ablations retain the same data partitions, optimizer family, temporal capacity, and downstream evaluation heads.

\subsection{Horizon-Wise Evaluation Metrics}
\label{sec:evaluation_metrics}

For horizon $k$, metrics are computed over the valid held-out context--target set
\begin{equation}
\Omega_k^{\mathrm{test}}
=
\left\{
(n,t)
\;\middle|\;
n\in\mathcal{D}_{\mathrm{test}},
\;
1\leq t\leq T-k
\right\}.
\end{equation}
Let
$\widehat{\mathbf{e}}_{n,t+k\mid t}$
and
$\mathbf{e}_{n,t+k}^{\star}$
denote the prediction and frozen SPEAR target.

\paragraph{Cosine similarity.}
Directional agreement is measured by
\begin{equation}
\begin{aligned}
\operatorname{CosSim}_{k}
&=
\frac{1}{|\Omega_k^{\mathrm{test}}|}
\sum_{(n,t)\in\Omega_k^{\mathrm{test}}}
\\[-0.5ex]
&\quad
\frac{
\widehat{\mathbf{e}}_{n,t+k\mid t}^{\top}
\mathbf{e}_{n,t+k}^{\star}
}{
\bigl(\lVert\widehat{\mathbf{e}}_{n,t+k\mid t}\rVert_2+\varepsilon\bigr)
\bigl(\lVert\mathbf{e}_{n,t+k}^{\star}\rVert_2+\varepsilon\bigr)
}.
\end{aligned}
\label{eq:metric_cosine}
\end{equation}

\paragraph{Mean squared and mean absolute error.}
Latent reconstruction errors are
\begin{align}
\operatorname{MSE}_{k}
&=
\frac{1}
{D|\Omega_k^{\mathrm{test}}|}
\sum_{(n,t)\in\Omega_k^{\mathrm{test}}}
\left\|
\widehat{\mathbf{e}}_{n,t+k\mid t}
-
\mathbf{e}_{n,t+k}^{\star}
\right\|_2^2,
\\
\operatorname{MAE}_{k}
&=
\frac{1}
{D|\Omega_k^{\mathrm{test}}|}
\sum_{(n,t)\in\Omega_k^{\mathrm{test}}}
\left\|
\widehat{\mathbf{e}}_{n,t+k\mid t}
-
\mathbf{e}_{n,t+k}^{\star}
\right\|_1.
\label{eq:metric_errors}
\end{align}

\paragraph{Coefficient of determination.}
To avoid high-variance latent dimensions dominating the score, $R^2$ is computed independently for each embedding coordinate and macro-averaged:
\begin{equation}
R_k^2
=
\frac{1}{D}
\sum_{d=1}^{D}
\left[
1
-
\frac{
\sum_{(n,t)\in\Omega_k^{\mathrm{test}}}
\left(
\widehat{e}_{n,t+k\mid t,d}
-
e_{n,t+k,d}^{\star}
\right)^2
}{
\sum_{(n,t)\in\Omega_k^{\mathrm{test}}}
\left(
e_{n,t+k,d}^{\star}
-
\overline{e}_{k,d}^{\star}
\right)^2
+
\varepsilon
}
\right],
\label{eq:metric_r2}
\end{equation}
where $\overline{e}_{k,d}^{\star}$ is the mean target value of coordinate $d$ over $\Omega_k^{\mathrm{test}}$.

For any horizon-specific metric $m_k$, the aggregate score is
\begin{equation}
\overline{m}_{w}
=
\sum_{k=1}^{K}
w_k m_k,
\label{eq:weighted_metric}
\end{equation}
using the same normalized horizon weights as temporal pretraining. Per-horizon curves are always reported alongside aggregate scores because weighted averages can conceal long-horizon failure.

\subsection{Horizon-Wise Forecasting Behaviour}
\label{sec:horizon_behaviour}

Forecasting quality is analyzed as a function of $k$ using
$\operatorname{CosSim}_{k}$,
$\operatorname{MSE}_{k}$,
$\operatorname{MAE}_{k}$, and
$R_k^2$.
Because the model predicts each horizon directly, longer-horizon degradation is not attributed to recursive error accumulation. Instead, it reflects increasing uncertainty, weaker dependence between the available context and distant targets, and the smaller number of valid context--target pairs available near the end of each sequence.

The horizon curves are compared with the persistence, linear-extrapolation, and seasonal-persistence baselines. Improvement over persistence indicates that the temporal encoder learns state evolution beyond local continuity, while improvement over seasonal persistence indicates that it captures context-dependent departures from a fixed annual cycle. Results should be reported over short, intermediate, and long horizon groups in addition to individual horizons.

\subsection{Embedding Magnitude Drift and Calibration}
\label{sec:magnitude_drift}

Cosine similarity can remain high even when predicted embeddings have incorrect scale. Two complementary statistics are therefore computed. Relative norm bias is defined as
\begin{equation}
\Delta_k^{\mathrm{norm}}
=
\frac{
\mathbb{E}_{\Omega_k^{\mathrm{test}}}
\left[
\left\|
\widehat{\mathbf{e}}_{n,t+k\mid t}
\right\|_2
\right]
-
\mathbb{E}_{\Omega_k^{\mathrm{test}}}
\left[
\left\|
\mathbf{e}_{n,t+k}^{\star}
\right\|_2
\right]
}{
\mathbb{E}_{\Omega_k^{\mathrm{test}}}
\left[
\left\|
\mathbf{e}_{n,t+k}^{\star}
\right\|_2
\right]
+
\varepsilon
}.
\label{eq:norm_bias}
\end{equation}
The normalized absolute norm error is
\begin{equation}
\operatorname{NormErr}_{k}
=
\mathbb{E}_{\Omega_k^{\mathrm{test}}}
\left[
\frac{
\left|
\left\|
\widehat{\mathbf{e}}_{n,t+k\mid t}
\right\|_2
-
\left\|
\mathbf{e}_{n,t+k}^{\star}
\right\|_2
\right|
}{
\left\|
\mathbf{e}_{n,t+k}^{\star}
\right\|_2
+
\varepsilon
}
\right].
\label{eq:norm_error}
\end{equation}

Post-hoc affine calibration is used only as a diagnostic of whether residual error is dominated by first- and second-order scale mismatch. For each horizon and embedding coordinate, calibration statistics are estimated on the validation set:
\begin{equation}
\widetilde{e}_{n,t+k\mid t,d}
=
\mu_{k,d}^{\star}
+
\sigma_{k,d}^{\star}
\frac{
\widehat{e}_{n,t+k\mid t,d}
-
\widehat{\mu}_{k,d}
}{
\widehat{\sigma}_{k,d}
+
\varepsilon
},
\label{eq:affine_calibration}
\end{equation}
where
$(\widehat{\mu}_{k,d},\widehat{\sigma}_{k,d})$
and
$(\mu_{k,d}^{\star},\sigma_{k,d}^{\star})$
are the validation-set moments of predicted and target embeddings, respectively. The fitted transformation is then applied unchanged to the test predictions. Improvements after calibration are reported as diagnostic evidence of scale or offset mismatch; calibrated results are not treated as the primary performance of the unmodified model.

\subsection{Temporal Self-Similarity Preservation}
\label{sec:self_similarity}

Seasonal and recurrent structure is examined through temporal self-similarity. The target self-similarity matrix is
\begin{equation}
S_{i,j}^{\star}
=
\frac{1}{N_{i,j}}
\sum_{n\in\mathcal{I}_{i,j}}
\frac{
\mathbf{e}_{n,i}^{\star\top}
\mathbf{e}_{n,j}^{\star}
}{
\left(
\left\|
\mathbf{e}_{n,i}^{\star}
\right\|_2+\varepsilon
\right)
\left(
\left\|
\mathbf{e}_{n,j}^{\star}
\right\|_2+\varepsilon
\right)
},
\label{eq:target_similarity}
\end{equation}
where $\mathcal{I}_{i,j}$ contains test sequences with valid states at both temporal positions and $N_{i,j}=|\mathcal{I}_{i,j}|$.

For a fixed horizon $k$, define the prediction associated with target position $j$ as
\begin{equation}
\widehat{\mathbf{e}}_{n,j}^{(k)}
=
\widehat{\mathbf{e}}_{n,j\mid j-k},
\qquad
j>k.
\end{equation}
The predicted self-similarity matrix is then
\begin{equation}
\widehat{S}_{i,j}^{(k)}
=
\mathbb{E}
\left[
\cos
\left(
\widehat{\mathbf{e}}_{n,i}^{(k)},
\widehat{\mathbf{e}}_{n,j}^{(k)}
\right)
\right].
\label{eq:predicted_similarity}
\end{equation}
Similarity-geometry preservation is summarized by the normalized Frobenius discrepancy
\begin{equation}
\mathcal{D}_{\mathrm{sim}}^{(k)}
=
\frac{
\left\|
\widehat{\mathbf{S}}^{(k)}
-
\mathbf{S}^{\star}
\right\|_{F}
}{
\left\|
\mathbf{S}^{\star}
\right\|_{F}
+
\varepsilon
},
\label{eq:similarity_discrepancy}
\end{equation}
computed over the temporal indices valid for horizon $k$. Heatmaps are inspected for diagonal continuity, repeated off-diagonal bands, and seasonal recurrence. Such structure is interpreted as temporal regularity in the learned latent space rather than as direct evidence of a specific biological process unless supported by corresponding labels or phenological observations.

\subsection{Prediction--Target Alignment}
\label{sec:alignment_heatmaps}

To determine whether the model predicts the correct future position rather than returning a temporally generic or identity-like embedding, prediction--target alignment matrices are computed:
\begin{equation}
H_{i,j}^{(k)}
=
\mathbb{E}_{n}
\left[
\cos
\left(
\widehat{\mathbf{e}}_{n,i+k\mid i},
\mathbf{e}_{n,j}^{\star}
\right)
\right].
\label{eq:alignment_matrix}
\end{equation}
A correctly aligned horizon-$k$ forecast should concentrate similarity near $j=i+k$. Diagonal concentration is quantified as
\begin{equation}
\operatorname{DC}_{k}
=
\frac{1}{|\mathcal{I}_{k}|}
\sum_{i\in\mathcal{I}_{k}}
\left[
H_{i,i+k}^{(k)}
-
\frac{1}{T-1}
\sum_{\substack{j=1\\j\neq i+k}}^{T}
H_{i,j}^{(k)}
\right],
\label{eq:diagonal_concentration}
\end{equation}
where $\mathcal{I}_{k}$ contains valid forecast origins. Positive and increasing diagonal concentration relative to the baselines indicates that the model has learned temporally specific progression rather than a degenerate constant or persistence solution.

\subsection{Loss and Temporal-Encoding Ablations}
\label{sec:ablation_analysis}

The composite objective is evaluated against cosine-only and regression-only variants. Cosine-only optimization tests whether directional alignment is sufficient, while regression-only optimization tests whether Euclidean reconstruction alone preserves latent geometry. The composite loss is assessed using the complete set of horizon-wise metrics, norm-drift statistics, and self-similarity discrepancy.

Temporal conditioning is evaluated using the following controlled variants:
\begin{enumerate}
    \item learned absolute position embeddings without calendar conditioning;
    \item RoPE without month--year embeddings;
    \item month--year embeddings without RoPE;
    \item combined RoPE and month--year conditioning;
    \item the complete model without QK normalization.
\end{enumerate}
RoPE and calendar conditioning are not treated as interchangeable. The former encodes ordering and relative displacement within the sequence, whereas the latter identifies seasonal phase and inter-annual context. The combined model is expected to be most useful when similar relative offsets correspond to different seasonal regimes.

The role of multi-horizon supervision is examined by comparing the complete model with a single-step model trained only at $k=1$ and with a uniformly weighted multi-horizon model. This determines whether long-range supervision improves the transferable representation or merely increases forecasting difficulty.

\subsection{Downstream Transfer and Representation Ablations}
\label{sec:downstream_transfer}

Because the abstract and central claims concern reusable Earth representations, embedding diagnostics are complemented by downstream transfer experiments on land-cover mapping and temporally demanding agricultural tasks. The same lightweight downstream architecture and data split are used for all representation variants. Three principal configurations are compared:
\begin{enumerate}
    \item \textbf{SPEAR:} frozen instantaneous spectral embeddings without temporal pretraining;
    \item \textbf{SPEAR + untrained temporal module:} the same temporal architecture with randomly initialized or task-trained temporal parameters, controlling for additional model capacity;
    \item \textbf{SPEAR-NeXT:} frozen or lightly adapted representations from the self-supervised temporal model.
\end{enumerate}
Where applicable, a prediction-head-only ablation is also included to separate the contribution of the pretrained temporal backbone from that of the downstream head.

Classification tasks are evaluated using overall accuracy and macro-averaged F1 score, with per-class results reported for imbalanced datasets. Regression tasks are evaluated using MAE, RMSE, $R^2$, and, where appropriate, percentage-based error. Geographic holdouts are used whenever the task is intended to assess cross-region transfer. Each reported gain is therefore interpreted relative to a capacity-matched baseline and not solely relative to a weaker downstream model.

Together, the forecasting diagnostics and downstream ablations test whether causal multi-horizon latent prediction produces representations that are not only predictable in embedding space but also more useful for Earth-observation transfer.

\section{Results and Discussion}
\label{sec:results_discussion}

This section reports the empirical behaviour of SPEAR-NeXT at three complementary levels: (i) future-state forecasting accuracy across temporal horizons, (ii) preservation of latent magnitude and temporal geometry, and (iii) transfer of the learned representations to downstream Earth-observation tasks. All forecasting results are computed on the spatially disjoint test partition described in Section~\ref{sec:dataset_construction}. Unless otherwise stated, the EMA model is used for evaluation, the SPEAR encoder remains frozen, and aggregate values use the normalized inverse-horizon weights in Eq.~\eqref{eq:weighted_metric}. Bracketed entries indicate values or figures to be inserted after the final experimental run.

\subsection{Downstream Representation Transfer}
\label{sec:downstream_results}

The practical value of SPEAR-NeXT is evaluated by transferring the learned
temporal representations to general land-cover mapping and temporally
demanding agricultural tasks. The India and CONUS models are pretrained
independently and are evaluated only on their corresponding regional
downstream datasets. Therefore, the India and CONUS results reported below
represent two separate regional learning pipelines rather than direct
cross-region transfer by a single model.

Where possible, the same downstream head, label split, and optimization
protocol are retained across representation models. This isolates
differences in pretrained representation quality from differences in
downstream model capacity.

\subsubsection{Regional Land-Cover Classification}
\label{sec:landcover_results}

Land-cover classification evaluates whether the temporally predictive
representations remain useful for general Earth-surface mapping rather than
only for agriculture-specific tasks. SPEAR-NeXT is compared with Presto
\cite{Tseng_2023_Presto} and Tessera \cite{Feng_2026_CVPR} using an MLP
classification head under the corresponding India and CONUS evaluation
protocols.

\begin{table*}[t]
\centering
\caption{
Land-cover classification using the independently pretrained India and
CONUS models. All methods use an MLP downstream head. Higher values are
better. Best results are shown in bold.
}
\label{tab:landcover_results}
\resizebox{0.88\textwidth}{!}{
\begin{tabular}{llcccc}
\toprule
& &
\multicolumn{2}{c}{India}
&
\multicolumn{2}{c}{CONUS} \\
\cmidrule(lr){3-4}
\cmidrule(lr){5-6}

Model
& Head
& Accuracy (\%) $\uparrow$
& F1 (\%) $\uparrow$
& Accuracy (\%) $\uparrow$
& F1 (\%) $\uparrow$ \\
\midrule

SPEAR-NeXT
& MLP
& \textbf{94.81}
& \textbf{94.79}
& \textbf{88.78}
& \textbf{90.66} \\

Presto \cite{Tseng_2023_Presto}
& MLP
& 85.20
& 84.87
& 86.17
& 86.61 \\

TESSERA \cite{Feng_2026_CVPR}
& MLP
& 72.35
& 74.23
& 86.40
& 88.53 \\

\bottomrule
\end{tabular}}
\end{table*}

On the India evaluation, SPEAR-NeXT achieves an accuracy of $94.81\%$ and
an F1 score of $94.79\%$. Relative to the strongest competing baseline,
Presto, these results correspond to gains of $9.61$ and $9.92$ percentage
points, respectively. The larger difference relative to Tessera suggests
that the India-pretrained SPEAR-NeXT representation captures temporal and
multimodal structure that is particularly useful under the India land-cover
distribution.

On the CONUS evaluation, SPEAR-NeXT obtains an accuracy of $88.78\%$ and an
F1 score of $90.66\%$. The corresponding gains over the strongest baseline,
Tessera, are $2.38$ percentage points in accuracy and $2.13$ percentage
points in F1 score. Although the margin is smaller than that observed in
India, SPEAR-NeXT remains the strongest model under both regional
evaluation settings. These results indicate that the benefit of predictive
temporal pretraining is not restricted to crop-specific tasks.

\subsubsection{SICKLE Multi-Task Agricultural Transfer}
\label{sec:sickle_results}

The India-pretrained model is further evaluated on the SICKLE benchmark
\cite{Sani_2024_WACV}, which includes crop classification and agricultural
regression tasks involving sowing date, harvest date, and yield as shown in table \ref{tab:sickle_multitask_results} . This
multi-task setting tests whether the representation captures not only crop
identity but also temporally structured agronomic information.

\begin{table*}[t]
\centering
\caption{
Multi-task transfer on SICKLE \cite{Sani_2024_WACV}. Higher is better for
binary crop-classification F1, while lower is better for MAPE. Best results
are shown in bold and second-best results are underlined.
}
\label{tab:sickle_multitask_results}
\resizebox{0.93\textwidth}{!}{
\begin{tabular}{lcccc}
\toprule
Model
& Binary classification F1 (\%) $\uparrow$
& Sowing-date MAPE (\%) $\downarrow$
& Harvest-date MAPE (\%) $\downarrow$
& Yield MAPE (\%) $\downarrow$ \\
\midrule

U-Net \cite{Ronneberger_2015_UNet}
& 87.71
& 2.34
& \underline{4.76}
& 72.38 \\

Time2Agri \cite{Gupta_2026_Time2Agri}
& 82.08
& \underline{1.58}
& 6.81
& 36.77 \\

Presto \cite{Tseng_2023_Presto}
& 89.87
& 5.05
& 5.93
& 23.30 \\

TESSERA \cite{Feng_2026_CVPR}
& \textbf{91.63}
& 4.13
& 7.73
& \underline{22.80} \\

SPEAR-NeXT
& \underline{90.30}
& \textbf{1.25}
& \textbf{3.03}
& \textbf{21.70} \\

\bottomrule
\end{tabular}}
\end{table*}

SPEAR-NeXT achieves the best performance on all three continuous
agricultural tasks. For sowing-date estimation, the MAPE decreases from
$1.58\%$ for the strongest baseline to $1.25\%$, corresponding to a
relative error reduction of approximately $20.9\%$. For harvest-date
estimation, SPEAR-NeXT reduces MAPE from $4.76\%$ to $3.03\%$, a relative
reduction of approximately $36.3\%$. Yield-prediction MAPE decreases from
$22.80\%$ for Tessera to $21.70\%$, corresponding to a relative reduction
of approximately $4.8\%$.

SPEAR-NeXT does not obtain the highest binary-classification score:
Tessera achieves an F1 score of $91.63\%$, compared with $90.30\%$ for
SPEAR-NeXT. The $1.33$ percentage-point difference indicates that the
proposed representation is not uniformly dominant across all downstream
objectives. However, its consistently stronger performance on sowing date,
harvest date, and yield suggests that causal multi-horizon pretraining is
particularly beneficial for tasks whose targets depend on temporal
progression rather than only on crop-category separability.

\subsubsection{Crop Classification under Fusion and Adaptation Strategies}
\label{sec:crop_classification_ablation}

\begin{table*}[t]
\centering
\small
\setlength{\tabcolsep}{4.5pt}
\renewcommand{\arraystretch}{1.15}

\caption{
\textbf{Crop-classification performance of SPEAR and SPEAR-NeXT under
different multimodal fusion, prediction-head, and adaptation strategies.}
All configurations are evaluated using $200$ labeled samples from five crop
classes. SPEAR denotes the previously published pixel-level spectral and
multimodal representation model~\cite{Ranjan_2026_WACV}.
\emph{Concat} denotes feature concatenation, whereas \emph{Cross} denotes
cross-attention-based feature fusion. \emph{MLP} denotes a multilayer
perceptron prediction head, \emph{RF} denotes a random forest prediction
head, and \emph{PEFT} denotes parameter-efficient fine-tuning.
\emph{Frozen} indicates that the pretrained representation encoder is kept
fixed during downstream training, whereas \emph{Unfrozen} permits
end-to-end parameter updating. The SPEAR macro-F1 values represent the
conservative lower bound of the observed approximately $3$--$5\%$ relative
difference from the corresponding SPEAR-NeXT results. Higher accuracy and
macro-F1 values indicate better classification performance.
}
\label{tab:crop_classification_adaptation}

\resizebox{\textwidth}{!}{%
\begin{tabular}{@{}lllccrrrr@{}}
\toprule
\multirow{2}{*}{Fusion}
& \multirow{2}{*}{Head}
& \multirow{2}{*}{Adaptation}
& \multirow{2}{*}{Samples}
& \multirow{2}{*}{Classes}
& \multicolumn{2}{c}{Accuracy (\%) $\uparrow$}
& \multicolumn{2}{c}{Macro-F1 $\uparrow$} \\
\cmidrule(lr){6-7}
\cmidrule(lr){8-9}
&
&
&
&
&
SPEAR
& SPEAR-NeXT
& SPEAR
& SPEAR-NeXT \\
\midrule

\multirow{6}{*}{Concat}
& \multirow{3}{*}{MLP}
& Frozen
& 200
& 5
& 77.56
& \textbf{83.33}
& 0.7865
& \textbf{0.8279} \\

&
& Unfrozen
& 200
& 5
& 80.77
& 82.05
& 0.7738
& 0.8145 \\

&
& PEFT
& 200
& 5
& 81.41
& 80.13
& 0.7593
& 0.7993 \\

\cmidrule(lr){2-9}

& \multirow{3}{*}{RF}
& Frozen
& 200
& 5
& 74.36
& \textbf{76.92}
& 0.7190
& \textbf{0.7568} \\

&
& Unfrozen
& 200
& 5
& 77.56
& 75.00
& 0.7029
& 0.7399 \\

&
& PEFT
& 200
& 5
& 76.28
& 74.36
& 0.6929
& 0.7294 \\

\midrule

\multirow{6}{*}{Cross}
& \multirow{3}{*}{MLP}
& Frozen
& 200
& 5
& 68.59
& 71.79
& 0.6464
& \textbf{0.6804} \\

&
& Unfrozen
& 200
& 5
& 63.46
& \textbf{80.77}
& 0.5976
& 0.6290 \\

&
& PEFT
& 200
& 5
& 51.28
& 78.85
& 0.4734
& 0.4983 \\

\cmidrule(lr){2-9}

& \multirow{3}{*}{RF}
& Frozen
& 200
& 5
& 63.46
& 66.03
& 0.5950
& 0.6263 \\

&
& Unfrozen
& 200
& 5
& 67.31
& 76.56
& 0.6324
& \textbf{0.6657} \\

&
& PEFT
& 200
& 5
& 56.41
& \textbf{78.85}
& 0.5292
& 0.5571 \\

\bottomrule
\end{tabular}%
}
\end{table*}

Table~\ref{tab:crop_classification_adaptation} shows that SPEAR-NeXT
outperforms SPEAR in nine of the twelve evaluated configurations. The best
overall performance is obtained using concatenation, an MLP head, and a
frozen encoder, reaching $83.33\%$ accuracy and $0.8279$ macro-F1. Its
$5.77$ percentage-point accuracy gain over SPEAR indicates that the
pretrained temporal representation transfers effectively without
end-to-end fine-tuning.

The largest improvements occur with cross-attention fusion. Under MLP--PEFT,
accuracy increases from $51.28\%$ to $78.85\%$, while RF--PEFT improves from
$56.41\%$ to $78.85\%$. In contrast, concatenation produces smaller and less
consistent gains. Overall, the results suggest that SPEAR-NeXT provides
strong frozen representations while also enabling more effective
higher-capacity multimodal fusion.

\begin{figure*}[t]
    \centering

    \begin{minipage}[t]{0.495\textwidth}
        \centering
        \includegraphics[
            width=\linewidth
        ]{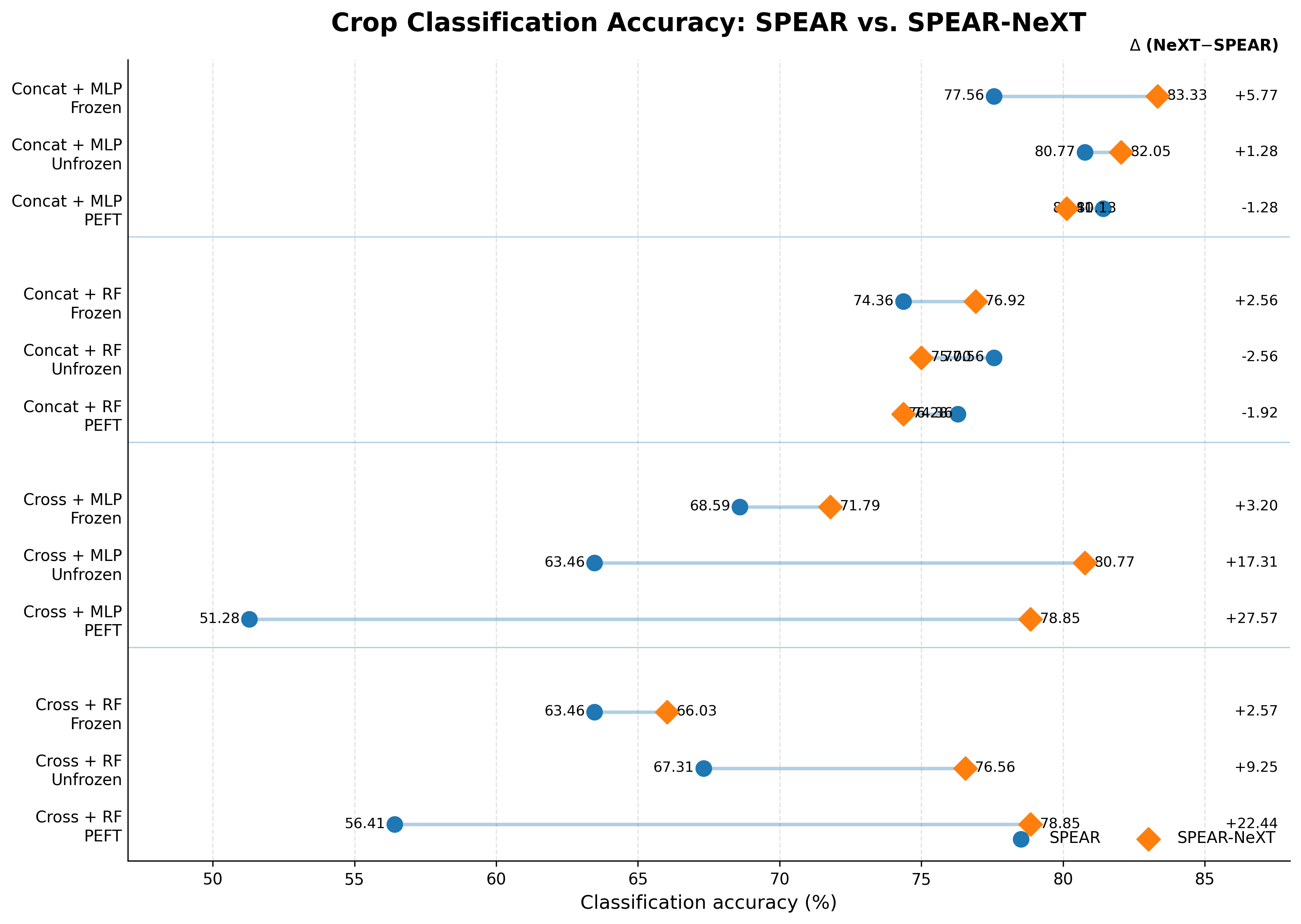}
        \vspace{-2mm}
        \textbf{(a) Classification accuracy}
    \end{minipage}
    \hfill
    \begin{minipage}[t]{0.495\textwidth}
        \centering
        \includegraphics[
            width=\linewidth
        ]{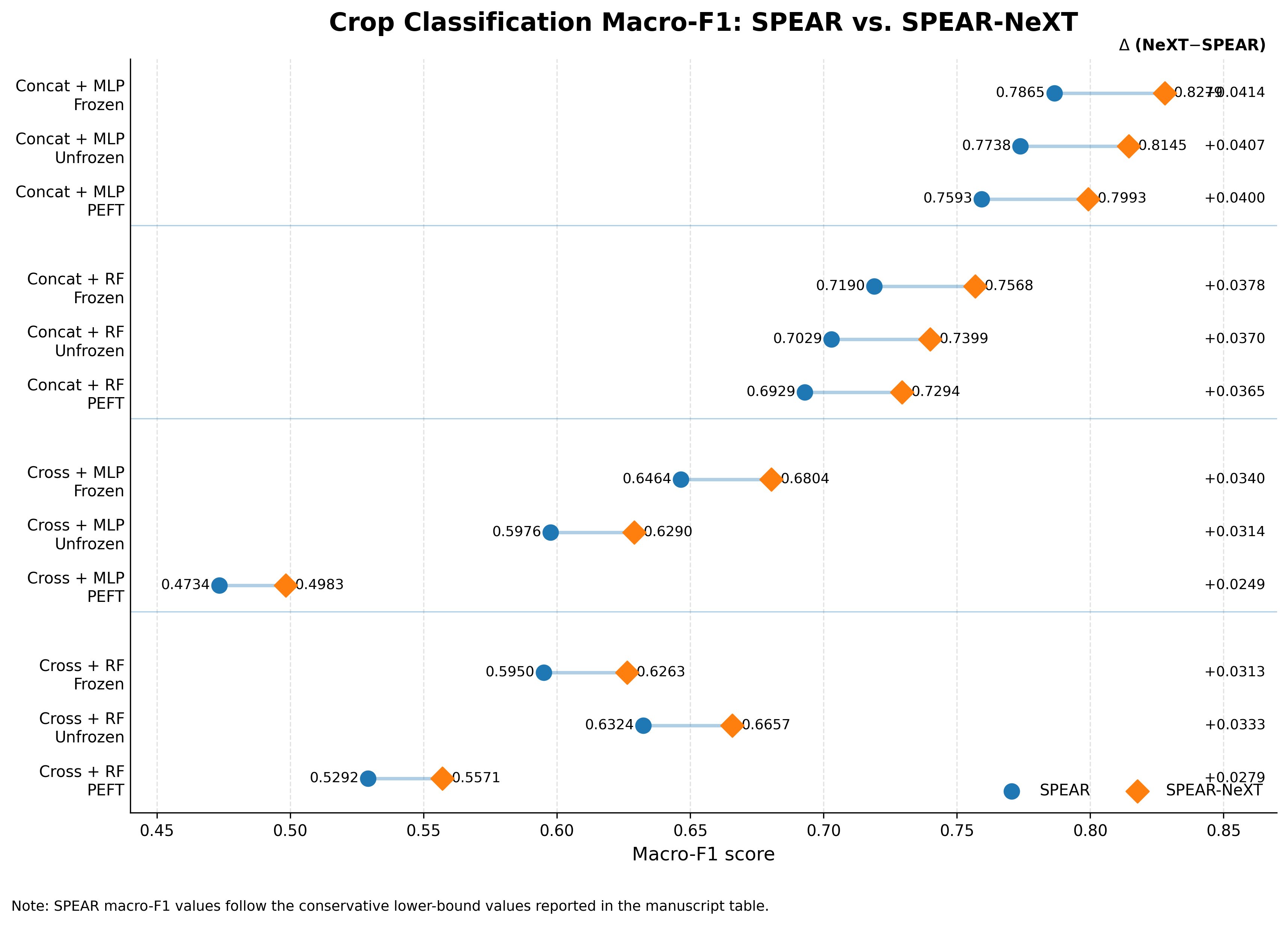}
        \vspace{-2mm}
        \textbf{(b) Macro-F1 score}
    \end{minipage}

    \caption{
    \textbf{Crop-classification performance of SPEAR and SPEAR-NeXT under
    different multimodal fusion, prediction-head, and adaptation settings.}
    Each horizontal pair compares SPEAR with SPEAR-NeXT for the same
    configuration, while the rightmost value reports the corresponding
    SPEAR-NeXT minus SPEAR difference. Concat denotes feature concatenation,
    Cross denotes cross-attention-based fusion, MLP denotes multilayer
    perceptron, RF denotes random forest, and PEFT denotes
    parameter-efficient fine-tuning. The SPEAR macro-F1 values correspond
    to the conservative lower-bound values reported in
    Table~\ref{tab:crop_classification_adaptation}.
    }
    \label{fig:crop_classification_transfer}
\end{figure*}

\subsubsection{USDA-NASS Crop-Yield Prediction}
\label{sec:nass_yield_results}

The CONUS-pretrained model is evaluated for crop-specific yield prediction
using USDA NASS data \cite{USDA_NASS_QuickStats}. Table
\ref{tab:nass_cropwise_results} reports the coefficient of determination
for the 2023 evaluation set. In the original presentation, the proposed
model was denoted as GeoSutra; it is referred to consistently here as
SPEAR-NeXT.

\begin{table}[t]
\centering
\caption{
Crop-specific USDA-NASS yield prediction for the 2023 evaluation year
\cite{USDA_NASS_QuickStats}. Values report $R^2$; higher is better.
}
\label{tab:nass_cropwise_results}
\resizebox{\columnwidth}{!}{
\begin{tabular}{lccc}
\toprule
Crop
& TESSERA \cite{Feng_2026_CVPR}
& Presto \cite{Tseng_2023_Presto}
& SPEAR-NeXT \\
\midrule
Corn    & 0.577 & 0.514 & \textbf{0.706} \\
Soybean & 0.704 & 0.718 & \textbf{0.803} \\
Wheat   & 0.778 & 0.754 & \textbf{0.831} \\
Cotton  & 0.517 & 0.465 & \textbf{0.542} \\
\midrule
Mean    & 0.644 & 0.613 & \textbf{0.721} \\
\bottomrule
\end{tabular}}
\end{table}

SPEAR-NeXT obtains the highest $R^2$ for every crop. Relative to the
strongest baseline for each crop, the absolute improvements are $0.129$ for
corn, $0.085$ for soybean, $0.053$ for wheat, and $0.025$ for cotton. The
mean crop-wise $R^2$ increases from $0.644$ for Tessera and $0.613$ for
Presto to $0.721$ for SPEAR-NeXT.

The largest improvement is observed for corn, while the smallest is
observed for cotton. This variation suggests that the value of temporal
representation learning depends on crop-specific signal quality, sample
availability, and the predictability of seasonal development from the
available satellite and environmental observations. The consistent
improvement across all four crops nevertheless indicates that the
CONUS-pretrained representation transfers beyond classification to
continuous agricultural prediction.

\paragraph{Contribution of the pretrained temporal representation.}

To determine whether the improvements in crop-yield prediction originate
from the self-supervised temporal representation or only from the
downstream prediction head, the complete SPEAR-NeXT pipeline is compared
with a prediction-head-only control. The full model uses the frozen SPEAR
state encoder, the pretrained SPEAR-NeXT temporal backbone, and the
crop-specific prediction head. The control retains the same prediction
head but excludes the pretrained temporal representation module. Both
configurations are evaluated using the same crop-specific data partitions,
forecast dates, and downstream optimization protocol.

\begin{figure*}[!t]
    \centering
    \includegraphics[
        width=0.96\textwidth,
        keepaspectratio
    ]{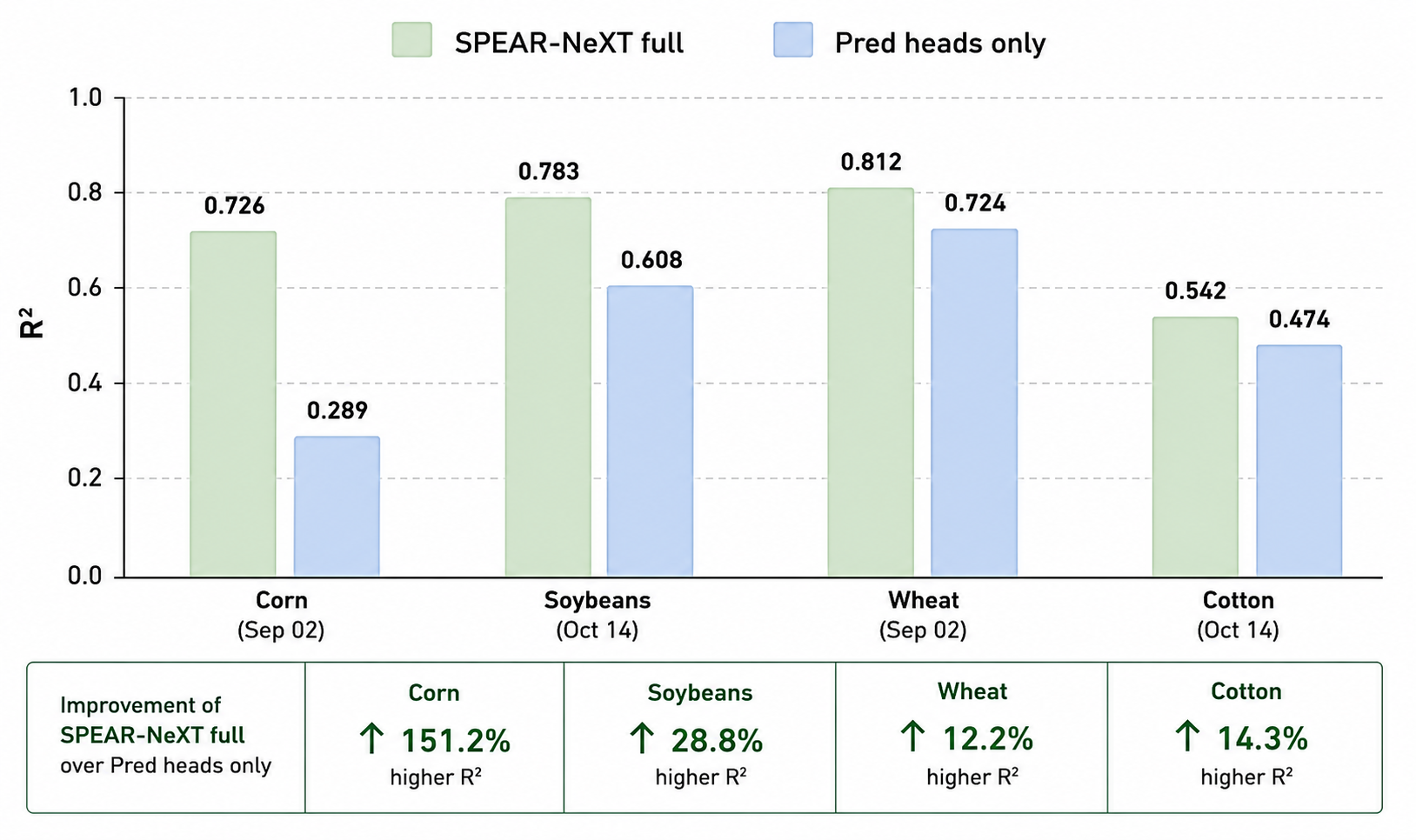}

    \caption{
    \textbf{Contribution of the pretrained SPEAR-NeXT temporal
    representation to crop-yield prediction.}
    Crop-specific $R^2$ values are compared between the complete
    SPEAR-NeXT pipeline and a prediction-head-only control at the indicated
    forecast dates. The complete model combines frozen SPEAR pixel states,
    the self-supervised SPEAR-NeXT temporal encoder, and the downstream
    prediction head, whereas the control excludes the pretrained temporal
    representation and uses only the prediction head. SPEAR-NeXT improves
    $R^2$ from $0.289$ to $0.726$ for corn, from $0.608$ to $0.783$ for
    soybean, from $0.724$ to $0.812$ for wheat, and from $0.474$ to $0.542$
    for cotton. These correspond to relative improvements of $151.2\%$,
    $28.8\%$, $12.2\%$, and $14.3\%$, respectively.
    }
    \label{fig:full_vs_prediction_head}
\end{figure*}

As shown in Figure~\ref{fig:full_vs_prediction_head}, the complete
SPEAR-NeXT model outperforms the prediction-head-only control for all four
crop types. The absolute $R^2$ improvements are $0.437$ for corn, $0.175$
for soybean, $0.088$ for wheat, and $0.068$ for cotton. The largest gain is
observed for corn, where the complete model increases $R^2$ from $0.289$ to
$0.726$. This substantial difference indicates that the downstream head
alone is insufficient to recover the temporal information required for
accurate yield prediction.

The improvements for soybean, wheat, and cotton are smaller but remain
consistent, showing that the benefit of the pretrained temporal
representation is not confined to one crop. Because the two configurations
use the same prediction-head architecture and evaluation protocol, the
performance difference can be attributed primarily to the representation
learned through self-supervised temporal pretraining rather than to
additional downstream-head capacity. These results provide direct evidence
that SPEAR-NeXT learns transferable temporal features that improve
crop-yield prediction beyond a task-specific prediction head.

\paragraph{Effect of spatial kriging.}

We further examine whether the crop-yield performance originates from the
learned representation or from the subsequent spatial interpolation stage.
The \emph{full model} consists of the frozen SPEAR encoder, the pretrained
SPEAR-NeXT temporal encoder, and the downstream yield-prediction head. The
\emph{prediction-head-only} configuration uses the same downstream head
without the pretrained SPEAR-NeXT temporal representation. A post-hoc
kriging stage is then applied independently to the predictions produced by
each configuration.

\begin{figure*}[!t]
    \centering
    \includegraphics[
        width=0.97\textwidth,
        keepaspectratio
    ]{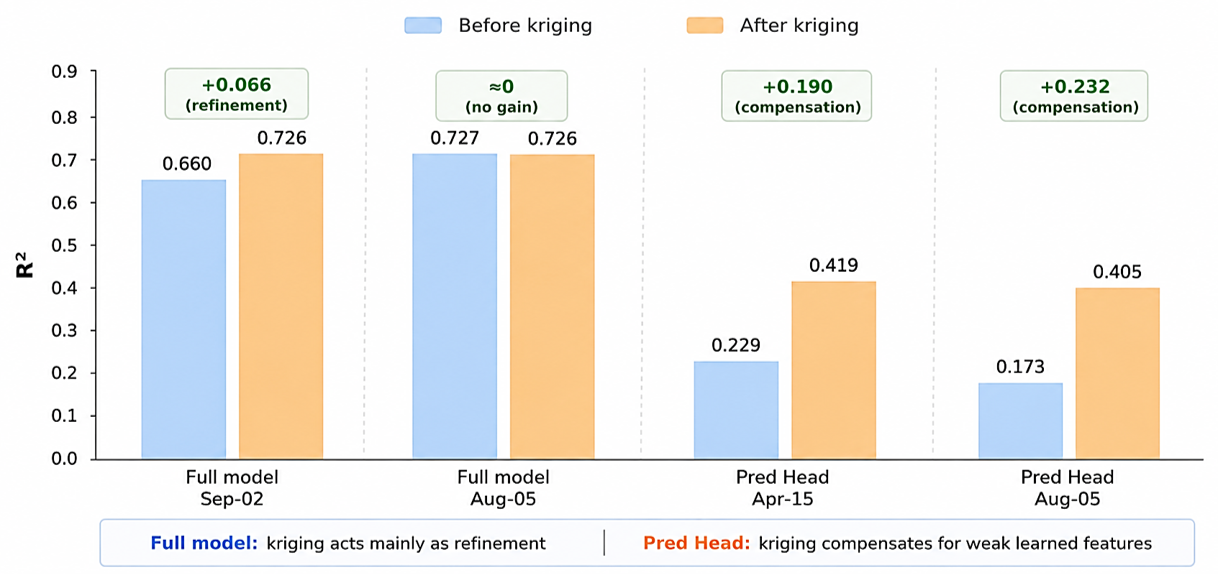}

    \caption{
    \textbf{Effect of post-hoc spatial kriging on corn-yield prediction
    using the complete SPEAR-NeXT model and the prediction-head-only
    control.}
    The complete model combines the frozen SPEAR encoder, the pretrained
    SPEAR-NeXT temporal encoder, and the downstream yield-prediction head.
    The prediction-head-only control excludes the pretrained temporal
    encoder and uses the same downstream head. Blue bars report performance
    before kriging, while orange bars report performance after spatial
    kriging. For the full model, kriging increases $R^2$ from $0.660$ to
    $0.726$ for the September~2 forecast and produces essentially no change
    for the August~5 forecast ($0.727$ to $0.726$). In contrast, the
    prediction-head-only configuration improves from $0.229$ to $0.419$ for
    the April~15 forecast and from $0.173$ to $0.405$ for the August~5
    forecast. These results indicate that kriging primarily refines the
    already informative SPEAR-NeXT predictions, whereas it compensates more
    strongly for the weaker spatial structure learned by the
    prediction-head-only model.
    }
    \label{fig:kriging_contribution}
\end{figure*}

Figure~\ref{fig:kriging_contribution} reveals a clear difference in how
spatial kriging affects the two model configurations. For the complete
SPEAR-NeXT model, the improvement is modest or negligible: $R^2$ increases
by $0.066$ for the September~2 forecast and remains effectively unchanged
for the August~5 forecast. This suggests that the pretrained spectral and
temporal representations already capture much of the spatially structured
signal relevant to corn-yield prediction, leaving kriging to perform only a
limited refinement.

The prediction-head-only control benefits substantially more from kriging.
Its $R^2$ increases by $0.190$ for the April~15 forecast and by $0.232$ for
the August~5 forecast. The larger post-processing gains indicate that the
head-only configuration produces weaker spatially coherent predictions and
therefore relies more heavily on spatial interpolation to recover regional
structure. Importantly, even after kriging, its performance remains below
that of the complete SPEAR-NeXT model. This supports the conclusion that
kriging cannot replace the information learned through self-supervised
temporal pretraining; it mainly acts as a complementary spatial refinement
stage.

\subsubsection{State-Wise In-Season Yield Forecasting}
\label{sec:statewise_yield_forecasting}

\begin{figure*}[!t]
    \centering
    \includegraphics[
        width=\textwidth,
        keepaspectratio
    ]{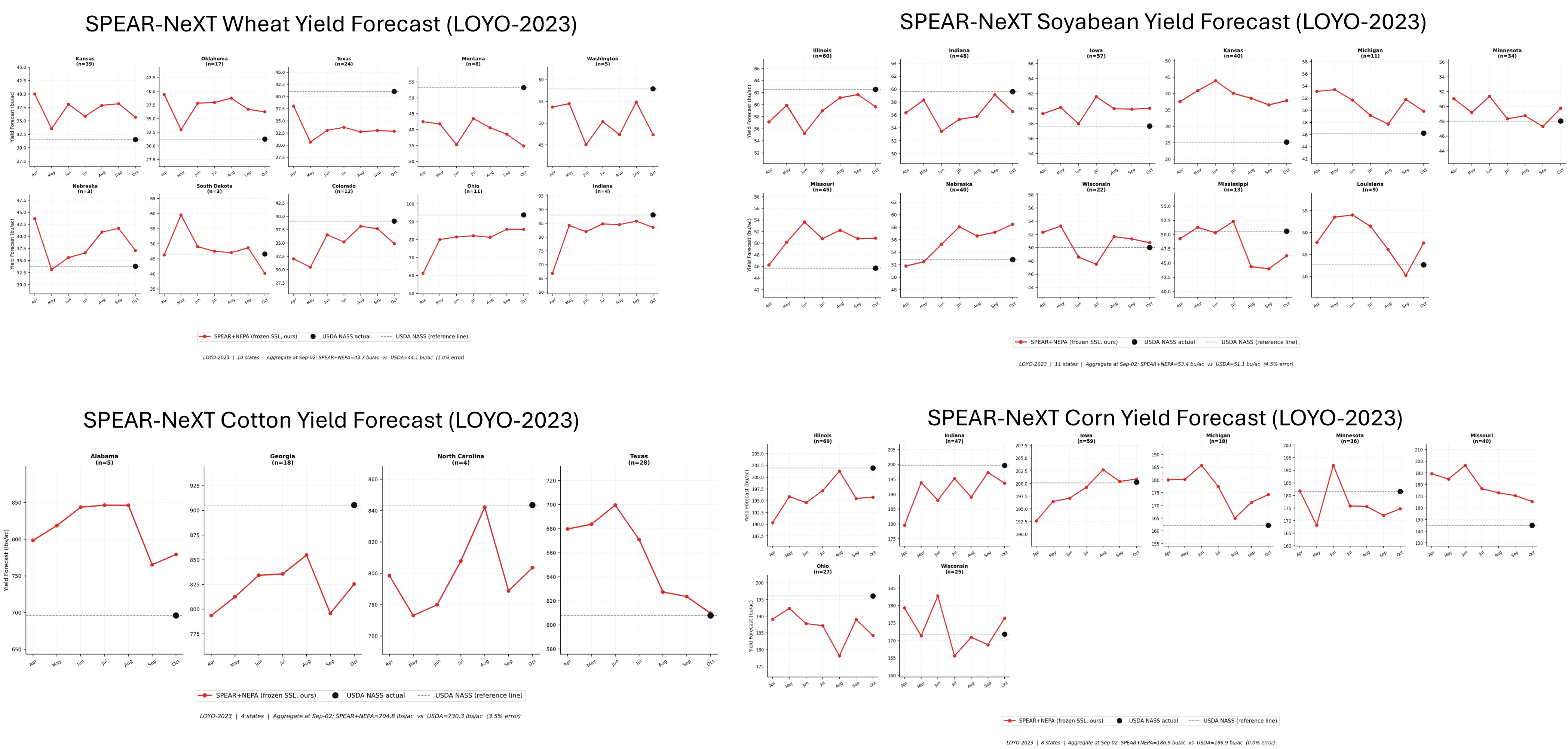}

    \caption{
    \textbf{State-wise in-season crop-yield forecasting under the
    leave-one-year-out 2023 evaluation.}
    Results are shown for
    \textbf{(a) wheat},
    \textbf{(b) soybean},
    \textbf{(c) cotton}, and
    \textbf{(d) corn}.
    Red trajectories denote SPEAR-NeXT yield predictions generated at
    successive in-season forecast dates, black markers indicate the final
    USDA NASS reported yields, and horizontal dashed lines provide the USDA
    NASS reference level for each state. The value $n$ shown in each panel
    denotes the number of spatial samples used for the corresponding state.
    The figure illustrates both the progressive updating of yield estimates
    during the growing season and the variation in forecasting difficulty
    across crops and geographic regions.
    }
    \label{fig:statewise_multicrop_yield}
\end{figure*}

Beyond the crop-level summary metrics, we examine how the predicted yield
changes across successive in-season forecast dates. Figure
\ref{fig:statewise_multicrop_yield} presents the leave-one-year-out
evaluation for 2023 across four major crops and their principal producing
states. Each red trajectory represents the sequence of SPEAR-NeXT yield
estimates obtained as additional observations become available during the
season. The black marker and horizontal dashed line denote the corresponding
final USDA NASS yield estimate \cite{USDA_NASS_QuickStats}.

Figure~\ref{fig:statewise_multicrop_yield} shows that SPEAR-NeXT responds
dynamically as additional within-season observations become available,
rather than producing a fixed seasonal estimate. For several states, the
forecast moves progressively toward the final USDA NASS value as the season
advances. The degree of convergence differs across crops and regions,
reflecting variation in crop calendars, environmental conditions, sample
availability, and the strength of the relationship between the observed
satellite trajectory and final yield.

The results also reveal cases in which early-season predictions deviate
substantially from the final reported yield before improving later in the
season. This behaviour is expected because early observations contain
limited information about late-season weather, stress, and management
effects. The state-wise analysis therefore complements the aggregate
crop-level $R^2$ results in Table~\ref{tab:nass_cropwise_results} by showing
when the model becomes informative during the season and where regional
forecasting uncertainty remains.
\subsubsection{Operational Comparison with NASA Acres}
\label{sec:nasa_acres_comparison}

To contextualize SPEAR-NeXT against an operational Earth-observation yield
forecasting system, we compare its 2024 corn-yield estimates with publicly
reported forecasts from NASA Acres. The NASA Acres system combines MODIS
Green Chlorophyll Vegetation Index, SERVIR Evaporative Stress Index,
AgERA5 precipitation, and a Generalized Additive Model to generate
county-scale corn and soybean yield forecasts across 12 Midwestern states.

\begin{figure*}[!t]
    \centering
    \includegraphics[
        width=0.98\textwidth,
        keepaspectratio
    ]{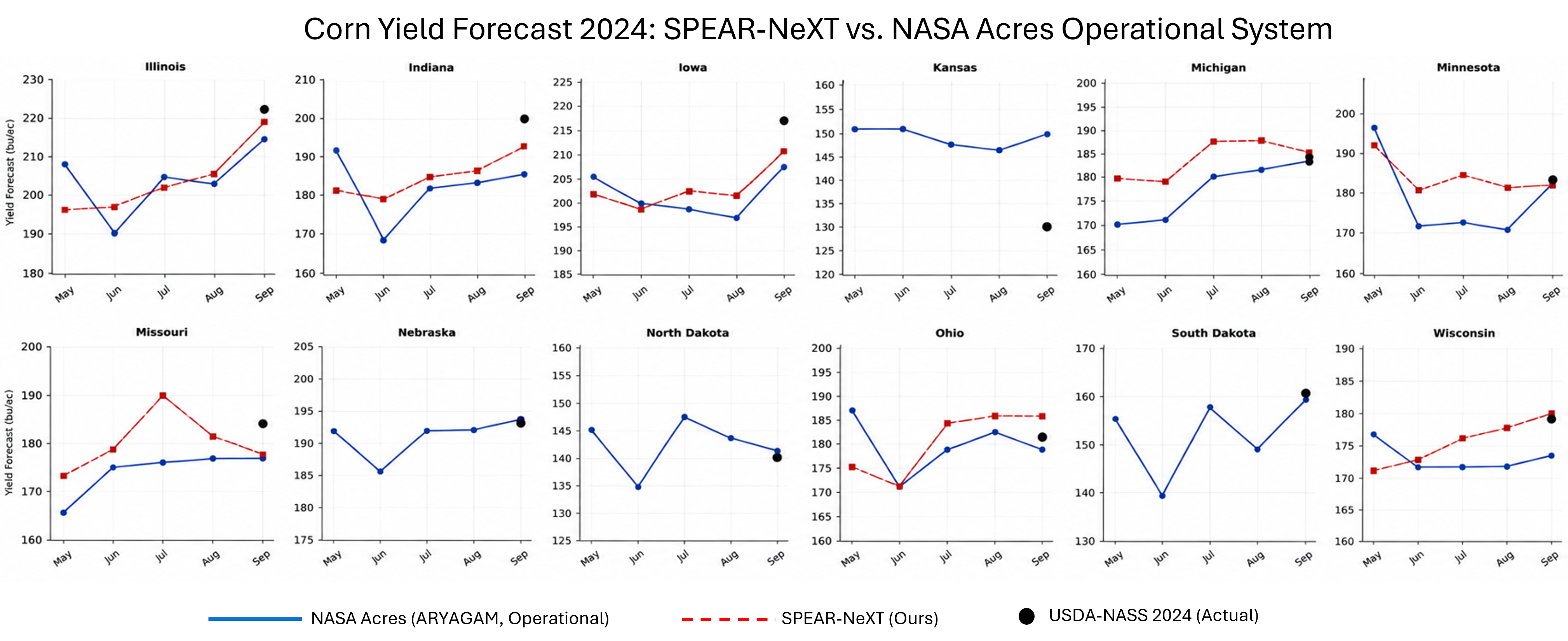}

    \caption{
    Comparison of 2024 corn-yield forecasts from SPEAR-NeXT and
    the NASA Acres operational forecasting system across 12 Midwestern
    states.
    Blue curves represent the publicly reported NASA Acres forecasts,
    red curves represent SPEAR-NeXT predictions, and black markers denote
    the final USDA NASS yield estimates. SPEAR-NeXT produces forecasts
    using the frozen self-supervised representation and a lightweight
    downstream yield-prediction head. This comparison is provided as an
    operational reference rather than a controlled benchmark because the
    two systems may differ in input data, spatial aggregation, forecast
    issue dates, preprocessing, and evaluation protocols.
    }
    \label{fig:nasa_acres_corn_comparison}

    \vspace{2pt}
    \begin{minipage}{0.98\textwidth}
        \footnotesize
        \textit{Source for the publicly reported NASA Acres forecasts:}
        \url{https://www.nasaacres.org/news/nasa-acres-supports-development-of-county-scale-yield-forecasting-tools}
    \end{minipage}
\end{figure*}

As shown in Figure~\ref{fig:nasa_acres_corn_comparison}, SPEAR-NeXT follows
the state-level variation in the final USDA NASS estimates while remaining
competitive with the NASA Acres operational forecasts. Across the 12-state
evaluation, SPEAR-NeXT obtains an aggregate forecast of
$191.8$~bu/ac against the corresponding USDA estimate of
$191.7$~bu/ac, with a state-level coefficient of determination of
$R^2=0.513$.

This comparison should be interpreted as an external operational reference
rather than as a controlled benchmark. The NASA Acres results are obtained
from publicly reported forecasts and may differ from the present experiment
in spatial aggregation, county weighting, input data, preprocessing,
forecast issue date, and the definition of the regional USDA reference.
Accordingly, the comparison demonstrates practical forecasting relevance
but is not used to claim direct superiority under an identical experimental
protocol.

\subsubsection{Summary of Downstream Transfer}
\label{sec:downstream_summary}

The downstream results reveal complementary strengths of SPEAR-NeXT.
Land-cover classification demonstrates transfer to general Earth-surface
mapping in both regional pipelines. SICKLE shows that the representation is
particularly effective for temporally structured agricultural regression,
although it does not outperform Tessera on binary crop classification.
Finally, the USDA-NASS experiment demonstrates consistent crop-wise gains
for yield prediction. Together, these findings support the use of
multi-horizon latent forecasting as a representation-learning objective
rather than only as an embedding-forecasting mechanism.

\subsection{Discussion}
\label{sec:discussion}

The results support the central distinction between instantaneous Earth-state
estimation and temporal state evolution. SPEAR provides compact multimodal
pixel representations from optical, radar, and environmental observations,
whereas SPEAR-NeXT learns how these latent states evolve under past-only,
causally masked multi-horizon prediction. The forecasting improvements over
persistence and seasonal-persistence baselines indicate that the temporal
module captures more than local continuity or a fixed annual cycle. The
prediction--target alignment and latent-geometry analyses further suggest
that the forecasts remain temporally specific rather than collapsing toward
an average future state.

The downstream results demonstrate that this predictive temporal objective
produces reusable representations. In regional land-cover classification,
SPEAR-NeXT achieves $94.81\%$ accuracy in India and $88.78\%$ in CONUS,
exceeding the strongest corresponding baselines by $9.61$ and $2.38$
percentage points, respectively. On SICKLE, SPEAR-NeXT obtains the lowest
errors for sowing-date, harvest-date, and yield prediction, although TESSERA
retains the highest binary-classification F1 score. This pattern suggests
that the main benefit of SPEAR-NeXT appears in tasks requiring phenological
and temporal reasoning rather than in every static classification setting.

The crop-classification adaptation study provides additional insight into
representation transfer. SPEAR-NeXT outperforms SPEAR in nine of twelve
configurations, with the best result obtained using concatenation, an MLP
head, and a frozen encoder, reaching $83.33\%$ accuracy and $0.8279$
macro-F1. The strong frozen-backbone result indicates that useful
crop-discriminative information is already encoded during temporal
pretraining. The largest gains occur with cross-attention under unfrozen and
PEFT adaptation, where SPEAR-NeXT substantially improves configurations
that perform poorly with SPEAR. Nevertheless, several concatenation-based
unfrozen and PEFT settings show small reductions, demonstrating that
additional downstream adaptation is not universally beneficial and may
disturb an already structured representation under limited supervision.

The crop-yield experiments provide the clearest evidence that the gains are
not attributable only to a stronger prediction head. SPEAR-NeXT achieves a
mean $R^2$ of $0.721$ across corn, soybean, wheat, and cotton, compared with
$0.644$ for TESSERA and $0.613$ for Presto. Improvements are observed for
all four crops, with the largest absolute gain occurring for corn. The
full-model versus prediction-head-only ablation further shows substantial
gains from the pretrained temporal representation, including $R^2$
improvements from $0.289$ to $0.726$ for corn and from $0.608$ to $0.783$
for soybean. These results indicate that the downstream head alone cannot
recover the temporal information learned during self-supervised
pretraining.

The kriging analysis leads to a similar conclusion. For the complete
SPEAR-NeXT model, spatial kriging provides only modest refinement or
negligible change, whereas the prediction-head-only configuration receives
much larger gains from kriging. Even after spatial interpolation, the
head-only model remains below the full model. Kriging therefore acts mainly
as a complementary spatial correction mechanism and does not replace the
temporally predictive representation learned by SPEAR-NeXT.

The objective and temporal-encoding ablations clarify the roles of the main
design components. Cosine alignment preserves directional structure in the
SPEAR latent space, while latent regression constrains coordinate scale and
reduces magnitude drift. Multi-horizon supervision exposes the encoder to
short-term continuity, intermediate transitions, and longer seasonal
dynamics, making it less dependent on single-step interpolation. RoPE and
calendar conditioning serve complementary roles: RoPE represents relative
order and temporal displacement, while month and year embeddings provide
seasonal phase and inter-annual context. Their combination is therefore
better suited to monthly Earth-observation sequences than either temporal
signal alone.

Several limitations remain. First, the prediction targets are frozen SPEAR
embeddings and consequently inherit the information limits and biases of
the underlying state encoder. Second, monthly compositing and
complete-coverage filtering simplify temporal learning relative to
operational satellite records containing clouds, missing acquisitions, and
irregular sampling. Extending the model with explicit elapsed-time
conditioning and missing-modality handling is therefore important. Third,
the India and CONUS models are pretrained and evaluated separately; the
reported results demonstrate regional transfer within each domain but do
not establish cross-region generalization. Fourth, the crop-classification
adaptation experiment uses only $200$ labeled samples and should be
validated across multiple sampling seeds and label budgets. Finally, high
latent-forecasting accuracy does not by itself guarantee physical
interpretability, and future work should relate individual latent
trajectories to measurable biophysical and agronomic processes.

Overall, the combined forecasting, representation, downstream-transfer,
prediction-head, and kriging analyses support causal multi-horizon latent
prediction as an effective self-supervised objective for compact
pixel-level Earth representations. SPEAR-NeXT is most valuable not because
it improves every downstream configuration uniformly, but because it
encodes temporal structure that transfers effectively through frozen or
lightweight adaptation to land-cover mapping, crop monitoring, phenological
estimation, and in-season yield forecasting.

{
    \small
    \bibliographystyle{ieeenat_fullname}
    \bibliography{main}
}

\clearpage
\appendix
%
%

\appendix
\section*{Appendix}
\label{sec:appendix}

\section{Dataset Information}
\label{app:dataset_information}

This appendix provides additional information about the datasets used to
pretrain the spectral and temporal components of SPEAR-NeXT and to evaluate
their downstream transfer. The experimental pipeline uses two
geographically distinct corpora. The India corpus supports the previously
published SPEAR spectral and multimodal pretraining stage
\cite{Ranjan_2026_WACV}, whereas the contiguous United States (CONUS)
corpus is used to construct the pixel-level temporal sequences required for
SPEAR-NeXT pretraining. Both corpora are sampled across heterogeneous
land-cover conditions using the Dynamic World taxonomy
\cite{Brown_2022_DynamicWorld}.


\begin{figure*}[!t]
    \centering

    \begin{subfigure}[t]{0.49\textwidth}
        \centering
        \includegraphics[
            height=5.2cm,
            width=\linewidth,
            keepaspectratio
        ]{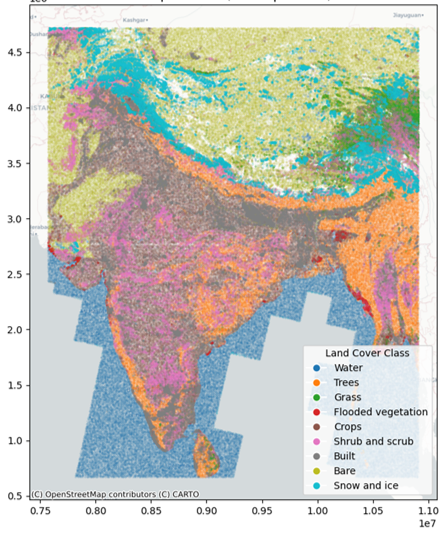}
        \caption{
        India pretraining corpus for SPEAR, containing approximately
        $2.7$ million pixel-level samples from Sentinel-2, PlanetScope,
        Sentinel-1, and matched climate--environmental variables.
        }
        \label{fig:appendix_india_dataset}
    \end{subfigure}
    \hfill
    \begin{subfigure}[t]{0.49\textwidth}
        \centering
        \includegraphics[
            height=5.2cm,
            width=\linewidth,
            keepaspectratio
        ]{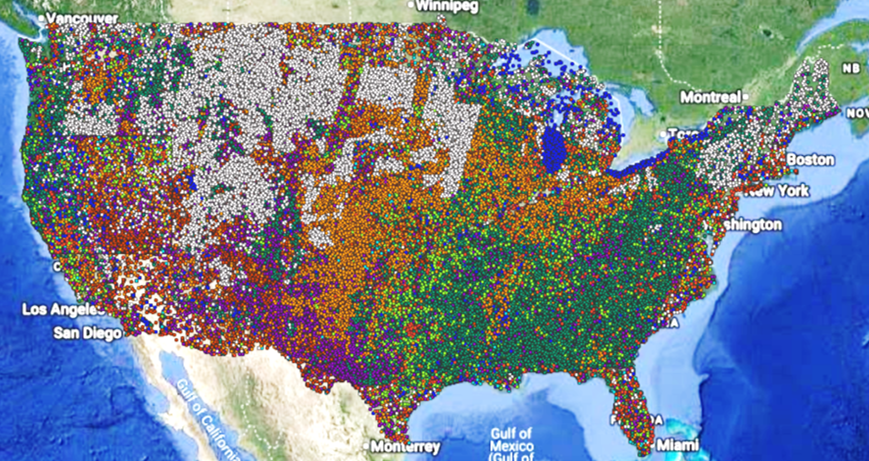}
        \caption{
        CONUS temporal-pretraining corpus for SPEAR-NeXT, containing
        approximately $4.5$ million pixel-level samples from Sentinel-2,
        Sentinel-1, and matched climate--environmental variables.
        }
        \label{fig:appendix_conus_dataset}
    \end{subfigure}

    \caption{
    \textbf{Geographic pretraining corpora and downstream evaluation
    datasets used in the SPEAR-NeXT pipeline.}
    The India corpus supports SPEAR spectral and multimodal pretraining
    \cite{Ranjan_2026_WACV}, while the CONUS corpus provides temporally
    ordered pixel sequences for SPEAR-NeXT pretraining. The associated
    downstream datasets include SICKLE \cite{Sani_2024_WACV},
    Sen1Floods11 \cite{Bonafilia_2020_CVPRW},
    VIIRS active-fire observations \cite{Schroeder_2014_VIIRS},
    USDA NASS Quick Stats \cite{USDA_NASS_QuickStats},
    CropHarvest \cite{Tseng_2021_CropHarvest}, and
    Dynamic World \cite{Brown_2022_DynamicWorld}.
    }
    \label{fig:appendix_dataset_overview}
\end{figure*}

Figure~\ref{fig:appendix_dataset_overview} summarizes the two geographically
distinct pretraining corpora and their associated downstream evaluation
tasks. The India corpus is shown in
Figure~\ref{fig:appendix_india_dataset}, while the CONUS corpus is shown in
Figure~\ref{fig:appendix_conus_dataset}.


\begin{table*}[!t]
\centering
\caption{
Summary of the pretraining corpora used in the SPEAR-NeXT pipeline.
}
\label{tab:appendix_pretraining_summary}
\resizebox{\textwidth}{!}{
\begin{tabular}{lllll}
\toprule
Corpus
& Approximate size
& Geographic coverage
& Input sources
& Primary role \\
\midrule

India
& $2.7$ million pixel-level samples
& India
& Sentinel-2, PlanetScope, Sentinel-1, climate/environmental variables
& SPEAR spectral and multimodal pretraining \\

CONUS
& $4.5$ million pixel-level samples
& Contiguous United States
& Sentinel-2, Sentinel-1, climate/environmental variables
& SPEAR-NeXT temporal pretraining \\

\bottomrule
\end{tabular}}
\end{table*}

\subsection{India Corpus}
\label{app:india_corpus}

\paragraph{Pretraining data.}

The India corpus contains approximately $2.7$ million pixel-level samples
distributed across diverse agricultural, forested, urban, barren, wetland,
and water-covered environments. This corpus was originally developed for
SPEAR and is described in detail in the corresponding publication
\cite{Ranjan_2026_WACV}. Sampling was stratified using Dynamic World
land-cover classes to reduce domination by highly prevalent surface types
and to increase coverage of less frequent classes
\cite{Brown_2022_DynamicWorld}.

The corpus combines the following observation sources:

\begin{itemize}

    \item \textbf{Sentinel-2 optical imagery.}
    Ten multispectral bands spanning visible, red-edge, near-infrared, and
    shortwave-infrared wavelengths are used. Sentinel-2 provides the
    principal multispectral input to the wavelength-aware SPEAR encoder
    \cite{Drusch_2012_Sentinel2}.

    \item \textbf{PlanetScope imagery.}
    Eight-band PlanetScope SuperDove observations introduce an additional
    optical sensor and a finer ground-sampling scale during multimodal
    SPEAR pretraining \cite{Planet_2026_PlanetScope}.

    \item \textbf{Sentinel-1 radar imagery.}
    Dual-polarization VV and VH synthetic-aperture-radar backscatter is
    represented at an approximate spatial resolution of $10$\,m
    \cite{Torres_2012_Sentinel1}.

    \item \textbf{Climate and environmental variables.}
    Daytime and nighttime land-surface temperature, precipitation, and
    elevation-related variables are matched to the sampled pixel
    locations. ERA5-Land provides the associated reanalysis information
    \cite{MunozSabater_2021_ERA5Land}. The complete feature composition
    and preprocessing follow the published SPEAR pipeline
    \cite{Ranjan_2026_WACV}.

\end{itemize}

\paragraph{Downstream evaluation.}

Representations learned from the India corpus are evaluated on several
complementary EO tasks:

\begin{itemize}

    \item \textbf{SICKLE crop classification.}
    SICKLE contains multisensor satellite time series from Landsat-8,
    Sentinel-1, and Sentinel-2, together with crop-type, phenological, and
    yield-related annotations collected over agricultural plots in the
    Cauvery Delta region of India \cite{Sani_2024_WACV}. The crop-type
    subset is used for agricultural classification.

    \item \textbf{Sen1Floods11 flood detection.}
    Sen1Floods11 contains georeferenced Sentinel-1 observations and
    corresponding surface-water labels spanning multiple flood events
    across several geographic regions \cite{Bonafilia_2020_CVPRW}. It is
    used to evaluate transfer to flood-related surface-water
    discrimination.

    \item \textbf{VIIRS fire-related classification.}
    The VIIRS $375$\,m active-fire product provides satellite-derived fire
    detections with improved spatial response relative to coarser fire
    products \cite{Schroeder_2014_VIIRS}. VIIRS detections are used as a
    reference source for the fire-related downstream evaluation.

    \item \textbf{Dynamic World land-cover classification.}
    Dynamic World provides near-real-time, $10$\,m land-cover predictions
    derived from Sentinel-2 imagery using a nine-class taxonomy
    \cite{Brown_2022_DynamicWorld}. These labels are used both for
    stratified sampling and general land-cover evaluation.

\end{itemize}

The India downstream suite spans agriculture, flooding, fire, and general
land-cover mapping. It therefore tests whether the SPEAR representation
transfers beyond a single semantic domain.

\subsection{CONUS Corpus}
\label{app:conus_corpus}

\paragraph{Temporal pretraining data.}

The CONUS corpus contains approximately $4.5$ million pixel-level samples
distributed across the contiguous United States. Its geographic extent
introduces substantial variation in vegetation, climate, agricultural
systems, management practices, and seasonal timing. Samples are stratified
using Dynamic World land-cover classes
\cite{Brown_2022_DynamicWorld}.

For SPEAR-NeXT, retained Sentinel-2 observations are chronologically
organized into pixel-level sequences of $T=61$ timesteps, as described in
Section~\ref{sec:dataset_construction}. Each Sentinel-2 observation is
transformed into a frozen $32$-dimensional SPEAR embedding before temporal
pretraining.

The broader CONUS corpus contains:

\begin{itemize}

    \item \textbf{Sentinel-2 optical imagery.}
    Ten multispectral bands are processed by the frozen SPEAR spectral
    encoder to produce a compact state at each timestep
    \cite{Drusch_2012_Sentinel2}.

    \item \textbf{Sentinel-1 radar imagery.}
    VV and VH radar backscatter observations are retained in the broader
    multimodal dataset construction and selected downstream evaluations
    \cite{Torres_2012_Sentinel1}.

    \item \textbf{Climate and environmental variables.}
    Matched temperature, precipitation, and elevation-related information
    is associated with the sampled locations. ERA5-Land provides the
    corresponding reanalysis fields
    \cite{MunozSabater_2021_ERA5Land}.

\end{itemize}

The temporal model presented in the main manuscript is pretrained on the
sequence of frozen Sentinel-2 SPEAR embeddings. Radar and climate variables
are part of the broader dataset and task-specific analyses, but are not
implicitly treated as temporal-encoder inputs unless explicitly stated in
the corresponding experiment.

\paragraph{Downstream evaluation.}

The CONUS experiments emphasize temporally demanding agricultural transfer
while retaining general environmental evaluation:

\begin{itemize}

    \item \textbf{USDA NASS crop-yield prediction.}
    County-level crop-yield records are obtained from the United States
    Department of Agriculture National Agricultural Statistics Service
    Quick Stats database \cite{USDA_NASS_QuickStats}. The database provides
    agricultural statistics indexed by commodity, geographic location, and
    year. These records are aligned with county-level satellite
    representations for yield prediction.

    \item \textbf{Sen1Floods11 flood detection.}
    Sen1Floods11 is used to evaluate transfer to flood-related surface-water
    discrimination \cite{Bonafilia_2020_CVPRW}.

    \item \textbf{CropHarvest crop/non-crop classification.}
    CropHarvest is a global, analysis-ready satellite dataset containing
    more than $90{,}000$ geographically diverse agricultural samples
    \cite{Tseng_2021_CropHarvest}. The binary crop/non-crop task is used to
    evaluate agricultural transfer under geographic variation.

    \item \textbf{Dynamic World land-cover classification.}
    Dynamic World labels are used for general nine-class land-cover
    evaluation \cite{Brown_2022_DynamicWorld}.

\end{itemize}

Relative to the India evaluation, the CONUS suite places greater emphasis
on temporal agricultural transfer through county-level yield prediction
and geographically distributed crop-presence classification.

\subsection{Role of the Datasets in the Experimental Design}
\label{app:dataset_roles}

Table~\ref{tab:appendix_dataset_roles} distinguishes datasets used for
self-supervised pretraining from those used only for downstream evaluation.
No downstream labels are used during SPEAR-NeXT temporal pretraining.

\begin{table*}[!h]
\centering
\caption{
Role of each data source in the SPEAR-NeXT experimental pipeline.
}
\label{tab:appendix_dataset_roles}
\resizebox{\textwidth}{!}{
\begin{tabular}{llll}
\toprule
Geographic setting
& Dataset or source
& Principal information
& Role \\
\midrule

India
& India multimodal corpus
& Optical, radar, and climate/environmental pixels
& SPEAR self-supervised pretraining \\

India
& SICKLE
& Multisensor agricultural time series and crop labels
& Crop-type classification \\

Global
& Sen1Floods11
& Sentinel-1 imagery and flood/surface-water labels
& Flood detection \\

Global
& VIIRS active-fire product
& $375$\,m active-fire detections
& Fire-related classification \\

India
& Dynamic World
& Nine-class Sentinel-2-derived land-cover labels
& Stratification and land-cover evaluation \\

\midrule

CONUS
& CONUS temporal corpus
& Pixel-level Sentinel-2 embedding sequences
& SPEAR-NeXT self-supervised temporal pretraining \\

CONUS
& USDA NASS Quick Stats
& County- and year-indexed crop-yield statistics
& Crop-yield prediction \\

Global
& CropHarvest
& Geographically diverse agricultural labels
& Binary crop/non-crop classification \\

CONUS / Global
& Dynamic World
& Nine-class Sentinel-2-derived land-cover labels
& Stratification and land-cover evaluation \\

\bottomrule
\end{tabular}}
\end{table*}

The combination of general mapping and agriculture-focused downstream
datasets is intended to determine whether SPEAR-NeXT learns reusable
temporal representations rather than features specialized for one region
or application. The India and CONUS corpora also provide an initial basis
for assessing transfer across contrasting land-cover distributions,
agro-climatic conditions, and seasonal regimes.


\end{document}